\documentclass[lettersize,journal]{IEEEtran}
\usepackage{amsmath,amsfonts}
\usepackage{algorithmic}
\usepackage{algorithm}
\usepackage{array}
\usepackage[caption=false,font=normalsize,labelfont=sf,textfont=sf]{subfig}
\usepackage{stfloats}
\usepackage{url}
\usepackage{verbatim}
\usepackage{graphicx}
\usepackage{cite}
\usepackage{hyperref}
\usepackage{stfloats}
\usepackage{lipsum}
\usepackage{comment}
\usepackage{amsfonts}
\usepackage{amsmath,amssymb,amsfonts}
\usepackage{url}
\usepackage{etoolbox}
\usepackage{graphicx}
\usepackage{optidef}
\usepackage{mathtools,amssymb}
\usepackage{ulem}
\usepackage{xcolor}
\usepackage{stix}
\usepackage{amsmath,mleftright,mathtools}
\usepackage[dvipsnames]{xcolor}

\usepackage[table,xcdraw]{xcolor}

\renewcommand{\textcolor}[2]{#2}

\usepackage{colortbl}
\usepackage{makecell}

\usepackage{subfig}
\usepackage{enumitem}
\usepackage{caption}

\usepackage{siunitx}

\DeclareMathOperator*{\argmin}{arg\,min}

\begin{document}

\title{
Optimal Safety-Aware Scheduling for Multi-Agent Aerial 3D Printing with Utility Maximization under Dependency Constraints
}

\author{Marios-Nektarios Stamatopoulos, Shridhar  Velhal, Avijit Banerjee, and George Nikolakopoulos\\
Robotics and AI Group, Luleå University of Technology, Sweden\\


}



\maketitle

\begin{abstract}
This article presents a novel coordination and task-planning framework to enable the simultaneous conflict-free collaboration of multiple unmanned aerial vehicles (UAVs) for aerial 3D printing. 
The proposed framework formulates an optimization problem that takes a construction mission divided into sub-tasks and a team of autonomous UAVs, along with limited volume and battery. 
It generates an optimal mission plan comprising task assignments and scheduling, while accounting for task dependencies arising from the geometric and structural requirements of the 3D design, inter-UAV safety constraints, material usage and total flight time of each UAV. The potential conflicts occurring during the simultaneous operation of the UAVs are addressed at a segment-level by dynamically selecting the starting time and location of each task to guarantee collision-free parallel execution.
An importance prioritization is proposed to accelerate the computation by guiding the solution towards more important tasks.
Additionally, a utility maximization formulation is proposed to dynamically determine the optimal number of UAVs required for a given mission, balancing the trade-off between minimizing makespan and the deployment of excess agents.
The proposed framework’s effectiveness is evaluated through a Gazebo-based simulation setup, where agents are coordinated by a mission control module allocating the printing tasks based on the generated optimal scheduling plan while remaining within the material and battery constraints of each UAV.
A video of the whole mission is available in the following link: \href{https://youtu.be/b4jwhkNPTyQ}{https://youtu.be/b4jwhkNPTyQ}.

\end{abstract}

\def\abstractname{Note to Practitioners}
\begin{abstract}
This framework addresses the critical need for efficiency and safety in planning and scheduling multiple aerial robots for parallel aerial 3D printing. Existing approaches lack safety guarantees for UAVs during parallel construction. This work tackles these challenges by ensuring safety during parallel operations and effectively managing task dependencies. The framework incorporates material and flight time constraints for each UAV and determines the optimal number of UAVs required for a specific construction mission in a single computation step. Additionally, a task prioritization method is introduced, significantly reducing the computational time of the optimization problem. This approach is particularly suited for applications such as rapid modular construction in remote or disaster-affected areas, where efficient UAV coordination is essential. Framework's preliminary feasibility is demonstrated in a simulated environment, while the real-world experimentation is planned as future work.

\end{abstract}


\section{Introduction}
\IEEEPARstart{R}{ecent} 
 advancements in additive manufacturing, combined with autonomous robotics, are revolutionizing next-generation construction and manufacturing techniques. Additive manufacturing has fundamentally reshaped the way complex production processes are approached, offering unmatched precision and efficiency. Its transformative impact is becoming increasingly apparent as the technology matures and is adopted across diverse industries \cite{reviewAddManAerospace,addManAutomotive,addManIndustry4.0}. Simultaneously, its potential for future innovations continues to grow. A particularly exciting development is its integration into the construction sector, leveraging automation to streamline construction processes, offer remarkable design flexibility, and enable the fabrication of intricate structures designed with Computer-Aided Design (CAD) tools \cite{addManCosntruction,craveiroa2019additive,addManConcreteCosntruction}.
\begin{figure}[t]
    \centering
    \includegraphics[width=\linewidth]{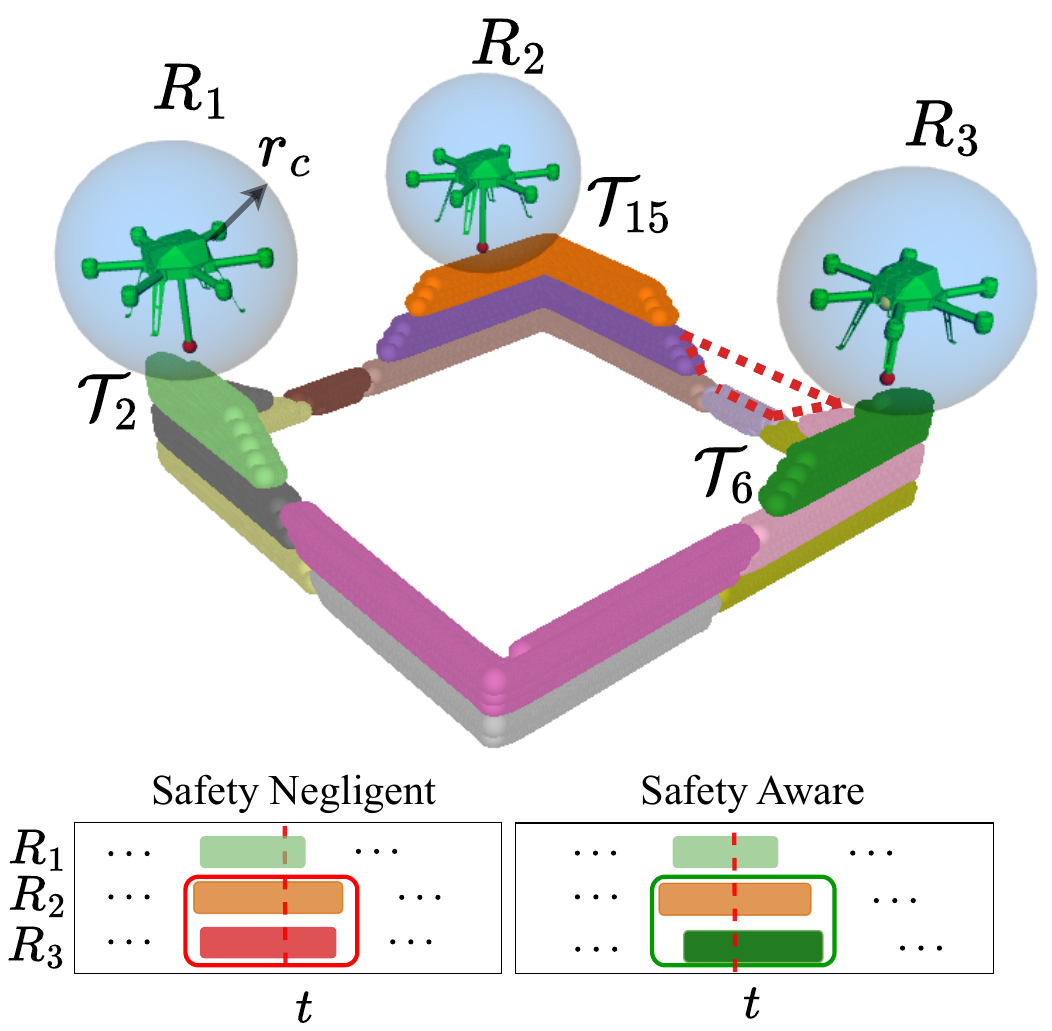}
    \caption{Concept figure of the proposed framework where three robots $R_1-R_3$ are printing the original mesh according to the optimal scheduling plan generated by the algorithm. Their minimum clearance $r_c$ is visualized with a light blue sphere. The mission schedule is generated in a way that all potential collisions between UAVs are resolved and safety is maintained. }
    \label{fig:conceptFig}
    \vspace{-5mm}
\end{figure}
Researchers have demonstrated the effective use of large-scale robotic arms and gantry systems to 3D print building components and even entire structures using various base materials \cite{additiveManConstruction2}. These methods, characterized by a high degree of automation, are especially advantageous for deployment in challenging environments, such as remote or inhospitable locations \cite{al2018large,bazli20233d}. For instance, these technologies can be critical in disaster-affected regions like those hit by earthquakes or wildfires, where the rapid construction of shelters and infrastructure is essential. Similarly, autonomous construction technologies hold immense promise for extraterrestrial habitats, where direct human intervention is minimal or absent \cite{mehaffy2017design}. However, the size and geometry of the workspace are often constrained by the scale of the equipment, complicating scalability. Mobile Additive Manufacturing (MAM) systems pose an emerging technology that could provide scalability for AM processes on construction sites through the co-operability of multiple mobile robots on individual 3D printing tasks \cite{addManufMobileRobotsChallenges}. Nevertheless, both these emerging robotic technologies and traditional construction methods face limitations, such as the need for pre-existing infrastructure to transport and assemble machinery on-site~\cite{xu2022robotics}.

Inspired by the natural construction methods of insects like bees and wasps, researchers are investigating cutting-edge advancements in robotics and Unmanned Aerial Vehicles (UAVs) for aerial additive manufacturing. This emerging concept, still in its early stages of development, presents a novel approach to construction \cite{kovac,zhang2022aerial,AUTCONstamatopoulos2025experiment}. Here, UAVs function as airborne builders, transporting and depositing construction material layer by layer to assemble structures. By introducing autonomous aerial systems into the construction sector, this innovative methodology has the potential to revolutionize how large-scale construction projects are executed, offering new opportunities for efficiency and automation. However, several hurdles need to be resolved to achieve a smooth implementation while employing multiple agents to operate simultaneously on the construction site. Major challenges include the restricted payload capacity of UAVs, their relatively short battery life compared to conventional construction equipment, and the intricate task of planning and managing autonomous operations while maintaining safety and coordination between them. 
Additionally, determining the number of agents that can effectively operate within a specified construction footprint remains a critical area of focus \cite{scienceRobAAMsurvey}.
Addressing these issues is essential to advancing the field.

\subsection{Related Work}

Multi-robot task assignment and scheduling have been studied in Operations Research (OR) and integrated process planning and scheduling (IPPS). However, research on using multiple aerial robots for construction automation remains limited, with most studies focusing on ground agents.
\subsubsection{Operations Research - Job-shop Scheduling Problem}
The particular case of the problem falls under the umbrella of the classic and challenging job-shop scheduling problem (JSSC) present in the field of OR and task allocation problems with temporal and ordering constraints \cite{taxonomyTaskAllocation}. It involves scheduling a set of jobs, each consisting of a sequence of tasks or operations, on a set of machines while minimizing the overall completion time and it has been widely researched over the years \cite{surveyJobScheduling,review2019jobshopIndustry4.0, TASEflexibjeJobShop}. 
In \cite{milpJobShopMobileRobots}, considering transport resources with mobile robots, a disjunctive graph model is modified to describe the relationship between the transport and processing tasks and a MILP model is proposed based on it.  Additionally, on energy-efficient scheduling \cite{flexibleJobShopmilpEnergy} proposed MILP models for flexible job shop scheduling to minimize energy consumption, showing improved efficiency. Researchers in \cite{JSCwithAGV} have also solved the JSSC problem  for autonomous ground vehicles (AGVs) using classic and quantum approaches, while in \cite{JSPdeadlcokMovemConsiderations}, 
position-based and network-based models are proposed to approach the JSC problem while taking into consideration possible deadlocks and the robots' movements. 

\subsubsection{Robotic Manufacturing-Construction}
The deployment of multiple robots for autonomous construction and manufacturing is often framed as a scheduling problem, focusing on coordinating tasks, resources, and timelines. 
Researchers have explored Deep Reinforcement Learning (DRL)-based methods for automated assembly planning in robot-driven prefabricated construction \cite{TASEprefabConstruction}. A runtime-aware scheduling approach using Petri nets is proposed for collaborative assembly tasks \cite{TASEpetriNetsScheduling}.
In multi-printer 3D manufacturing, tasks are scheduled by dividing products into parts with multiple printing options, optimized using a mathematical model and genetic algorithm \cite{scheduling3DPrinters}.
For large aircraft structures, multi-robot task scheduling with precedence constraints employs an auction-based method and machine learning to select efficient heuristics \cite{spacecraftmanufacBalanced}.
Task allocation for robots manufacturing complex aerospace structures aims to balance workloads and avoid collisions through precedence orders, using a hybrid of construction heuristics and iterated local search \cite{schedulingAircraftSoftPrecedence}.
In another study \cite{wenchaoZhouScheduling}, \textcolor{blue}{makespan} minimization and collision-free scheduling of multiple ground robots 3D printing simultaneously is investigated by splitting the mesh into tasks using a predefined way and defining dependencies between them. Using a dynamic dependency list algorithm, solutions that could result in collisions are eliminated and sequences of printing tasks are generated greedily based on the number of the available resources, while a genetic algorithm is also proposed to generate various printing schedules.

\textcolor{blue}{
Existing multi-robot 3D printing frameworks mainly handle conflicts through spatial or temporal exclusion. Ground-based systems such as \cite{wenchao_zhou_ground_robots} typically partition the workspace or enforce sequential exclusion, operating at the task or segment level with heuristic or rule-based exclusion. Collision checks, when used, are static feasibility tests that classify assignments as valid or invalid, without supporting temporal flexibility or adaptive coordination. As a result, these methods prevent conflicts through proactive isolation rather than dynamic resolution, limiting parallel efficiency.
Dynamic approaches, such as \cite{youwasps}, enable robots to pause and replan trajectories to avoid collisions. While effective for general coordination, such interruptions disrupt the continuous deposition required for uniform aerial printing. Similarly, \cite{3dPrintingTeamRobots} decouples planning into independent path computation and coordination-diagram-based retiming stages. Although computationally efficient, this separation treats collision avoidance as post-processing and cannot guarantee feasibility during execution. Moreover, retiming alters velocities, degrading print quality in tasks requiring constant motion.
}
\textcolor{blue}{
Additionally, in the aforementioned works, scheduling and task allocation are treated independently from collision management and conflict resolution.
Tasks are typically allocated first and subsequently filtered or corrected for conflicts, resulting in a decoupled process where assignments are not guaranteed to be conflict-free and lack awareness of motion-planning constraints. This separation hinders the optimal coordination, assignment, and scheduling of multi-robot printing operations.
}

In the specific case of aerial 3D printing, the multi-agent implementation is still in its infancy. In the current state of the art \cite{nature_aerial_AM}, a real-life implementation of an aerial worker manufacturing on a small scale is presented. In the emulated multi-agent scenario, the mesh is sliced into layers and each layer is further split into print jobs available to the robots. Constraints related to task manufacturability are enforced, and each robot selects its next task in real-time based on proximity to it.
On 
other
work \cite{stamatopoulosICUAS}, a centralized module is reactively assigning the printing tasks to the available UAVs with respect to the manufacturability dependencies and is based on employing heuristic probabilistic scores, while at the same time tasks with the most dependencies are prioritized first, reducing the complexity of the overall mission.
It must be noted that in 
other
works, task execution was primarily sequential. Multiple agents operated simultaneously only in \cite{stamatopoulosICUAS} so far, but task assignment and conflict resolution were managed reactively, without formal safety guarantees or a constant printing speed.

Existing solutions to safety-aware multi-task scheduling problems often address safety conflicts by ignoring them during planning and handling them reactively during execution, either by one of the UAVs pausing in place \cite{stamatopoulosICUAS} or halting the print and moving away\cite{nature_aerial_AM}. These heuristic approaches do not provide any guarantees and impose the risk of significantly altering the plan if the resolution proves infeasible \cite{nature_aerial_AM, stamatopoulosICUAS} or end in a deadlock, making planning non-deterministic and the mission unpredictable.
Other approaches enforce hard constraints to prevent conflicts where neighboring tasks with potential conflicts are disallowed from being executed simultaneously \cite{spacecraftmanufacBalanced, wenchaoZhouScheduling}. 
\textcolor{blue}{}
These methods reduce the flexibility of the planner, as the added constraints hinder the optimal allocation and scheduling of tasks, ultimately undermining the quality of the final mission plan.
These approaches do not directly address the safety-aware collaborative task scheduling problem. Thus, the need for a framework emerges to compute multi-task UAV schedules that minimize 3D construction makespan while ensuring collision-free simultaneous operation.

\subsection{Contributions} 
In this article, we propose an optimal scheduling and task allocation framework that generates a collision-free construction plan for UAVs, minimizing the total duration of the whole mission while considering the material and flight time budget of each UAV, along with the dependencies of the tasks due to their geometrical features. 
The potential collisions between UAVs executing printing tasks in a co-working environment are predicted and resolved while scheduling agents to tasks and the computation is accelerated by giving importance to the tasks with higher dependencies. Furthermore, the proposed method computes the number of aerial agents and their schedule to maximize the utility of the team. 
The simulation results validate the effectiveness of the proposed method for computing the safety-aware scheduling of aerial agents for additive construction. A faster solution can be obtained by slightly compromising on makespan minimization.
The key contributions of the proposed article can be listed as follows,
 \begin{itemize}
     \item  Safety-aware optimal scheduling for the whole mission with guarantee on the collision-free scheduling to minimize the overall mission duration, while following the dependencies of the geometry of construction, battery, and material limitations.
    \item  An importance prioritization is proposed to accelerate the computation by guiding the solution towards more important tasks 
    \item One-step optimization is proposed to compute the optimal team size of aerial agents and their schedules to maximize the agent utility.
 \end{itemize}
To the best of the authors' knowledge, this is the first time in the literature that a safety-aware, multi-task scheduling approach has been proposed for multi-agent aerial 3D printing.
\\

The rest of this paper is organized as follows. The  architecture of the Multi-Agent
Aerial Additive Construction framework, along with the notation of the article, is presented in Section \ref{sec:problemFormulation}. The formulation of the overall scheduling optimization problem, along with different variations of it and its integration into the overall mission, is explained in Section \ref{sec:methodology}. The scheduling results for different cases and variations of the proposed optimization problem, along with a case study simulation of the whole mission, are presented in Section \ref{sec:results}, while the article is concluded in Section \ref{sec:conclusion}.

\section{Problem Formulation} \label{sec:problemFormulation}

\subsection{Multi-Agent
Aerial Additive Construction Architecture}
\begin{figure*}
    \centering
    \includegraphics[width=0.98\linewidth]{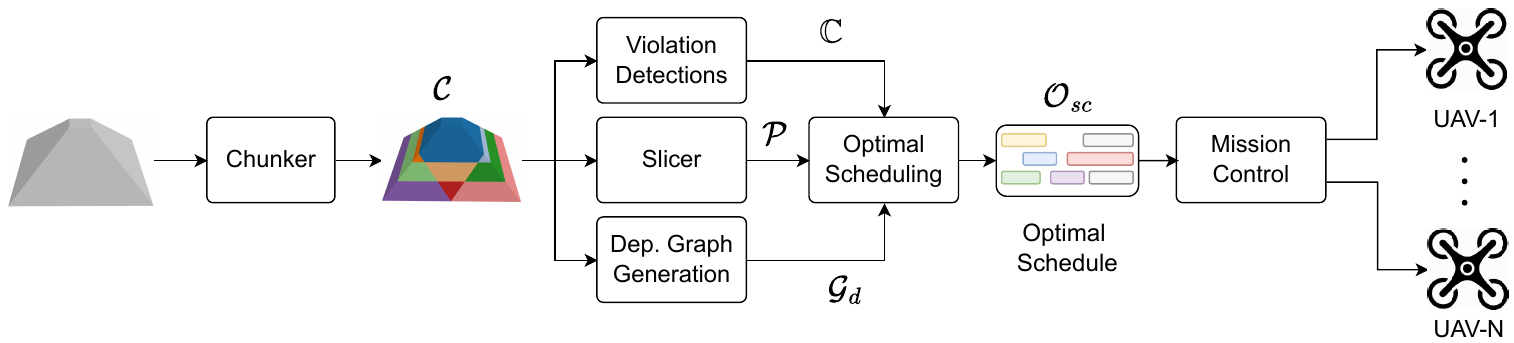}
    \caption{Multi-Agent
Aerial Additive Construction System Architecture Overview }
    \label{fig:blockDiagram}
\end{figure*}

A schematic overview of the overall architecture, showcasing the different blocks of the overall framework, is shown in Fig. \ref{fig:blockDiagram}. Initially, the desired mesh $\mathcal{M}$ to be constructed is fed to the chunker block \cite{stamatopoulosChunkingJINT} that decomposes it into smaller sub-parts referred to as 'chunks' $\mathcal{C}$ corresponding to the tasks $\mathcal{T}_i$ of the construction mission. The inter-dependencies between the chunks are identified and systematically placed in a graph $G_d$ via the dependency graph generator \cite{stamatopoulosICUAS}. All chunks are transformed into manufacturing paths $\mathcal{P}$ that need to be traversed by the UAVs in order to construct them through a slicer module according to \cite{StamatopoulosEmulation}. The paths are evaluated in pairs and all the clearance violations between their segments are systematically detected and placed in the set $\mathbb{C}$. 
All manufacturing paths, their interdependencies, and segment violations, along with the configuration of the available construction UAVs ($\mathcal{C}_{UAV}$), are provided as input to the optimal scheduler module. In this module, an optimization problem is formulated and solved to plan and coordinate the overall mission effectively.
The output of the module is a matrix $O_{sc}$ specifying the starting time of each task $\mathcal{T}_i$ and assigning it to a UAV $R_k$ corresponding to the scheduling for the whole construction mission. Finally, the matrix is fed to the mission control, which communicates and coordinates all UAVs in real-time, ensuring that the execution is carried out according to the scheduled plan.

\subsection{Scheduling Problem}

In this paper, we focus on the computation of the multi-task scheduling of UAVs for printing the chunks while satisfying the dependency constraints and guaranteeing safety during the execution of tasks.
Let us consider tasks of printing N chunks, denoted by 
{
$\mathcal{T} = \{\mathcal{T}_0, \mathcal{T}_1, ... , \mathcal{T}_{N-1} \}$. 
}
Each task  $\mathcal{T}_i$ requires $v^T_i$ volume of material for printing. The printing duration required for  $\mathcal{T}_i$ is denoted by $d_i$. The time at which task $\mathcal{T}_i$'s execution starts is denoted by $S_i$. The task $\mathcal{T}_i$ completed at completion time is denoted by $C_i$ where $C_i = S_i + d_i$. 
\\
Let us also consider a fleet of $M$ UAV robots, denoted by 
{
$\mathcal{R} = \{R_0, R_1, ..., R_{M-1} \}$}. A robot $R_k$ has a limited volume carrying capability denoted by  $v^R_k$. The  volume capacity of robots is denoted as set $\mathcal{V}^R = \{v^R_0, v^R_1, ..., v^R_{M-1}\}$.
Additionally, to ensure safety, UAVs operating in the same environment need to be separated from each other. The minimum distance required between any two UAVs for safety is denoted by $r_{c}$.  
\\
To achieve optimal print quality, it is crucial to ensure uniform material deposition throughout the printing process. Therefore, the UAVs must maintain a constant speed while printing to ensure a high-quality result.
In this paper, we assume that UAVs are homogeneous and operate at a constant speed $V_{ex}$ while printing. An overview of all the variables used throughout the formulation of the optimal scheduling problem, along with their description, is shown in Table \ref{tab:variables}.
\\
The tasks are generated by dividing the structure into smaller chunks. These tasks are both geometrically and structurally interconnected, meaning they have precedence constraints that dictate the order in which they must be executed. For example, the top layer of chunks cannot be constructed before completing the bottom layer. After the chunking phase, the dependencies between tasks are identified and represented using a dependency graph $G_d$ \cite{stamatopoulosICUAS}, which captures the feasible orders of execution.
\\
The UAVs must be scheduled to tasks while satisfying the dependency constraints imposed by $G_d$. In a co-working environment, however, UAVs may face potential collisions that prevent them from executing tasks simultaneously. To ensure safety, a minimum distance must be maintained between the UAVs at all times.
While the dependency graph imposes static constraints that remain fixed, the collision constraints are dynamic and depend on the UAVs’ schedules. Therefore, the scheduling problem must address both the static constraints from the dependency graph and the dynamic collision constraints among the UAVs.

\begin{table}[]
\centering
\resizebox{0.95\linewidth}{!}{ 
    \begin{tabular}{cc}
    \hline
        {Variable} & {Description} \\ \hline
        $\mathcal{T}_i$ & $i$-th Printing task \\
        $R_k$ & $k$-th Robot \\
        $v^R_k$ & Material budget of robot $R_k$ \\
        $d_i$ & Duration of task $\mathcal{T}_i$ \\
        $v^T_i$ & Volume of task $\mathcal{T}_i$ \\
        $x_{i,k}$ & Assignment of task $\mathcal{T}_i$ to  $R_k$ \\
        $S_i$ & Start time of task $\mathcal{T}_i$ \\
        $y^k_{i,j}$ & Order between $\mathcal{T}_i, \mathcal{T}_j$ assigned to $R_k$ \\
        $t^\kappa_i$ & Arrival time at the end of $i$-th segment of path $P_\kappa$ \\
        $\tau^{log}_s$ & Logistics time before start of each task \\
        $\tau^{log}_e$ & Logistics time after end of each task \\
        $t^F_k$ & Total flight time of robot $R_k$ \\
        $t^b_k$ & Remaining battery flight time of robot $R_k$ \\
        $C_{max}$ & Mission makespan \\
        $\alpha_i$     & Importance factor of task $\mathcal{T}_i$\\
        $u_k$     & Utilization of robot $R_k$\\
        \hline
    \end{tabular}
}
\caption{Variables used for the formulation of the Optimal Scheduling Problem.}
\label{tab:variables}
\end{table}

\section{Safety-aware Multi-task Scheduling Framework}\label{sec:methodology}

\subsection{Base Problem Formulation} \label{sec:constraints}
The task scheduling algorithm needs to be in accordance with typical task assignment 
constraints and the precedence constraints defined by the dependency relation of the chunks and encapsulated in the set $\mathcal{P}$.

Initially, the starting time of all tasks cannot be a negative number since the mission starts at $t=0$; thus, the following constraint is added.

\subsubsection{Task Assignment}
Initially, given the assumption that the mission starts at time $t=0$, all starting times $S_i$ must be non-negative:
\begin{equation}
    \label{eq:startTime}
    S_i \geq 0, \forall \mathcal{T}_i \in \mathcal{T}
\end{equation}
The assignment of tasks to the available robots is handled via the matrix $\mathbf{X} \in \mathbb{B}^2$ where each binary element $x_{i,k} = 1$ if the task  $\mathcal{T}_i$ is assigned to the robot $R_k$ and $0$ otherwise. Each task $\mathcal{T}_i$ can be assigned to only one robot $R_k$, and this rule is handled by the classic constraint
\begin{equation}
    \label{eq:assigment}
    \sum^{M-1}_{k=0} x_{i,k} = 1, \forall \mathcal{T}_i \in \mathcal{T}
\end{equation}

\subsubsection{Precedence}
\begin{figure*}
    \centering
    \includegraphics[width=\linewidth]{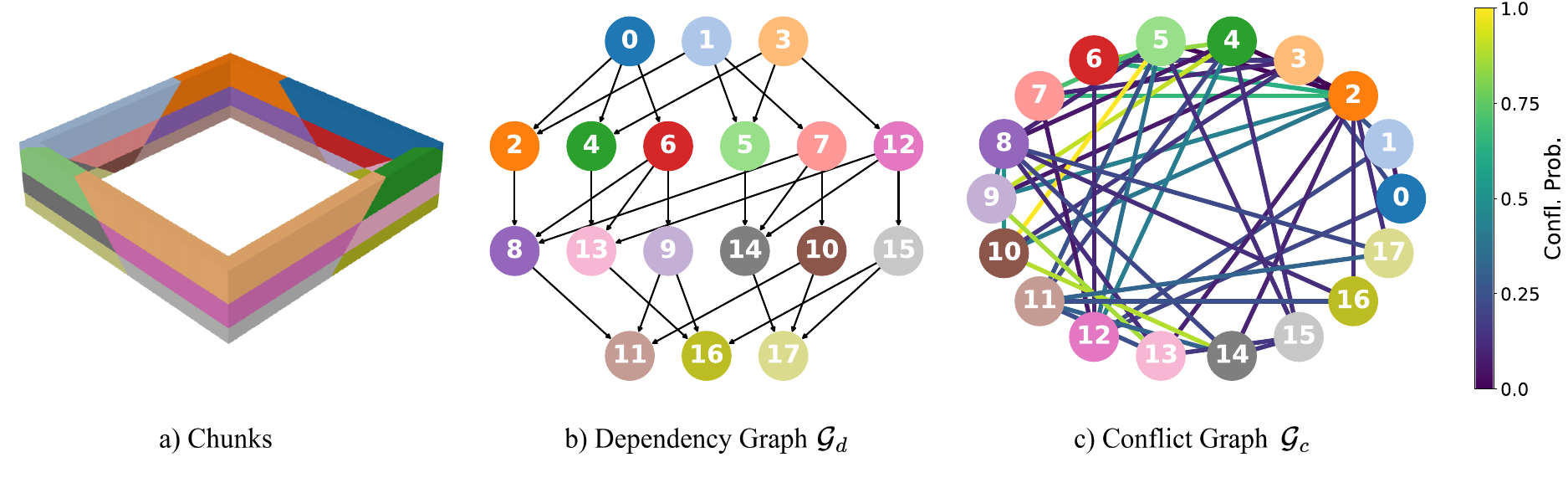}
    \caption{Rectangular mesh $M$ decomposed in $18$ chunks in a color-coded format (a) along with its dependency graph $G^d$ (b) and the Conflict Probability Graph $\mathcal{G}_c$ for pairs of tasks (c).}
    \label{fig:chunksFeatures}
    \vspace{-5mm}
\end{figure*}

The dependency relations between the printing tasks for the original mesh are captured in a precedence constraints set, denoted as 
%
\begin{equation}
{
\mathcal{P}_{pr} = \left\{ (\mathcal{T}_i, \mathcal{T}_j) \in \mathcal{T} \times \mathcal{T} \,\middle|\, (\mathcal{T}_i, \mathcal{T}_j) \in \mathcal{G}_d \right\}
}
\end{equation}
Each tuple $(\mathcal{T}_i, \mathcal{T}_j)$ in this set represents that task $\mathcal{T}_i$ must be completed before task $\mathcal{T}_j$ can begin.
This set $\mathcal{P}_{pr}$ is derived from the dependency graph $\mathcal{G}^d$, where every edge $(\mathcal{T}_i, \mathcal{T}_j) \in \mathcal{G}^d$ (as depicted in Fig. \ref{fig:chunksFeatures}.b) indicates that the manufacture of chunk $\mathcal{C}_i$ needs to be completed before  $\mathcal{C}_j$, implying  that the start time $S_j$ of task $\mathcal{T}_j$ needs to be selected after the end time $C_i$ of task $\mathcal{T}_i$.

By imposing these precedence relations and leveraging the fact that the duration $d_i$ of each task $\mathcal{T}_i$ is known, assuming the UAV maintains a constant speed while traversing the paths, the dependencies between the chunks are formally expressed as precedence constraints as follows:
\begin{equation}
    \label{eq:precedence}
    S_i + d_i \leq S_j, \forall (\mathcal{T}_i,\mathcal{T}_j) \in \mathcal{P}_{pr}
\end{equation}

\subsubsection{Ordering}
For each robot $R_k$, the order of the tasks being assigned to it is handled by the binary variable $y^k_{i,j}$ denoting that out of the two tasks assigned to robot $R_k$, the task $\mathcal{T}_i$ precedes $\mathcal{T}_j$ and is defined as follows:
\begin{equation}
    \label{eq:ordering1}
    y^k_{i,j} =  
    \begin{cases} 
        1 & ,\text{if } S_i + d_i \leq S_j  \\
        0 & ,\text{otherwise }
    \end{cases}
\end{equation}

This piecewise relation is transformed into linear constraints using the big-M method \cite{bigM_notation} 
\begin{equation}
    S_i + d_i \leq S_j + O(1 - y^k_{i,j}),\: \forall \mathcal{T}_i \neq \mathcal{T}_j \in \mathcal{T}, R_k \in \mathcal{R}
\end{equation}
where $O \in \mathbb{R}$ is a number big enough such as $O > d_i$, $\forall d_i \in \mathcal{D}$

In case two tasks $\mathcal{T}_i, \mathcal{T}_j$ are assigned to a robot $R_k$, the order between them can be defined by only one  of the binary variables $y^k_{i,j}$
since having both of them equal to $1$ would lead to ambiguity. Additionally, a robot can not execute two assigned tasks simultaneously, which is a fact that leads to the constraint expressed by the following equation:
\begin{equation}
    \label{eq:ordering2}
    y^k_{i,j} + y^k_{j,i} = x_{i,k} \  x_{j,k},\: \forall \mathcal{T}_i \neq \mathcal{T}_j \in \mathcal{T}, R_k \in \mathcal{R}
\end{equation}
indicating that in case both tasks are assigned to the robot $R_k$ ($x_{i,k} \  x_{j,k}=1$), only one ordering variable will be equal to $1$ as well. 

\subsubsection{Material Budget}
Each robot $R_k$ is equipped with a material capacity of $v^R_k$, representing the volume of material it carries. To ensure feasibility, the total volume of material required for all tasks $\mathcal{T}_i$ assigned to robot $R_k$ must not exceed this capacity. This constraint can be expressed as:

\begin{equation}
    \label{eq:materialBudget}
    \sum_{i=0}^{N-1} v^T_{i} x_{i,k} \leq v^R_{k}, \forall R_k \in \mathcal{R}
\end{equation}

where $x_{i,j}$ denotes whether task $\mathcal{T}_i$ is assigned to robot $R_k$ and $v^T_i$ represents the material volume required by task $\mathcal{T}_i$. This constraint guarantees that each robot operates within its material budget.

\subsubsection{Battery Life - Flight Time }
Given the fact that each robot has a limited battery charge and there may be cases where, despite a robot carrying sufficient material, its battery is depleted at some point and is rendered unable to continue the printing process. In order to incorporate handling of the aforementioned cases during the planning of the tasks, a flight time estimate $t^b_k \in \mathbb{R}$ is associated with every robot $R_k$. This estimate can be based on the battery voltage of each robot \cite{batteryLifeSina}, which is sampled at the start of the mission, but this estimation method is left out of the scope of this work.
 For the calculation of the actual time each UAV is airborne, the duration $d_i$ of each assigned task $\mathcal{T}_i$ to it is considered along with the logistics times $\tau^{log}_s$ and $\tau^{log}_e$ incorporating the motion of the UAV from its current position to the starting point of the task and backwards.

Following to that, the total flight time $t^F_k \in \mathbb{R}$ for a UAV $R_k$ is calculated as follows:
\begin{equation}
    t^F_k = \sum_i^{N-1} (\tau^{log}_s + d_{i} + \tau^{log}_e) x_{i,k}, \;R_k \in \mathcal{R}
\end{equation}
and consequently, it is constrained with an upper bound equal to the estimate of the remaining flight time $t^b_i$ based on the battery percentage.
\begin{equation}
    \label{eq:timeBudget}
    t^F_k  \leq t^b_k, \; \forall R_k \in \mathcal{R}
\end{equation}
\subsection{Safety-Aware Scheduling}\label{sec:conflictFreeConstraints}
During the simultaneous printing execution of multiple neighboring chunks, there is a chance of conflicts appearing between the agents due to their working envelopes, despite that their respective paths do not intersect. Thus, a minimum clearance $r_c \in \mathbb{R}$ between any pair of UAVs is imposed, indicating the minimum allowed distance the UAVs will have between them during execution.
{
Each manufacturing path $P_i$ associated with task $\mathcal{T}_i$ is represented as a sequence of linear segments $P_i = \{s^i_0, s^i_1, \dots, s^i_{N_i -1 }\}$, where $s^i_\kappa$ denotes the $\kappa$-th segment and $N_i$ is the total number of segments in path $P_i$. 
The probability $p_{i,j}$ of a conflict arising from the simultaneous execution of paths $P_i$ and $P_j$ is defined as:
}
\begin{equation}
    {
    p_{i,j} = \frac{\sum_{\kappa=0}^{N_i - 1} \sum_{\lambda=0}^{N_j-1} \mathrm{v}^{i,j}_{\kappa, \lambda}}{N_i N_j}
    }
\end{equation}

{
where
}

\begin{equation}
    {
    \mathrm{v}^{i,j}_{\kappa,\lambda} = 
    \begin{cases}
        1, & \text{if } d_{\min}(s^i_\kappa, s^j_\lambda) \leq r_c \\
        0, & \text{otherwise}
    \end{cases}
    }
\end{equation}
is a conflict indicator for the segment pair $(s^i_\kappa, s^j_\lambda)$, set to 1 when the minimum distance between the pair of segments  $d_{\min}(s^i_\kappa, s^j_\lambda) \leq r_c$ is less than the clearance $r_c$.
The pairwise conflict probabilities $p_{i,j}$ are systematically evaluated for all concurrently executed task pairs and are illustrated in Fig.~\ref{fig:chunksFeatures}.c in the form of a conflict graph $\mathcal{G}_c$. In this graph, nodes represent tasks, and edges indicate a non-zero probability of conflict. The color of each edge corresponds to the conflict probability, as defined by the accompanying color map.

Additionally, during the printing execution of each chunk, it is noticed that the optimal deposition of the material in layers is conducted when the UAV is tracking the manufacturing paths at a constant fixed speed $V_{ex}$, which is selected after manual iterations and experimentation. Thus, in the current work, aiming to enhance the printing performance, the extra requirement of keeping the speed of the UAVs constant and equal to $V_{ex}$ during printing operation is considered. This further restricts the flexibility of solving the problem and other approaches where the speed of the UAVs was modulated within the area of the fixed speed $V_{ex}$ \cite{stamatopoulos2024conflict}, can not be employed.
So, the handling of the conflicts between the printing UAVs is conducted by dynamically selecting the starting time $S_i$ of each task, which directly associates the position of the UAV during the execution of task $\mathcal{T}_i$ given the fact that the speed remains constant. The resolution scheme is not carried out in two distinct steps and is not decoupled from the scheduling process compared to other approaches \cite{nature_aerial_AM, stamatopoulosICUAS}, but is incorporated in the overall mission plan calculation step, enabling the optimal selection of the starting time. Specifically, the conflict resolution mechanism is formulated as additional constraints to the overall optimization problem of the scheduling as follows:

\subsubsection{Conflicting Segments}
Initially, in order to identify these conflicts, all segments $s^\kappa_i$ of manufacturing paths are evaluated in pairs, where $s^\kappa_i$ denotes the $i$-th segment of the path ${P}_\kappa \in \mathcal{P}$. Specifically, a conflicting set $\mathbb{C}$ is calculated as follows:
\begin{equation}
    \mathbb{C} = \{  (s^\kappa_i,s^\lambda_j) : d_{min}(s^\kappa_i,s^\lambda_j) \leq r_{c}, \forall (P_\kappa,P_\lambda) \in \binom{\mathcal{P}}{2}\}
\end{equation}
where a conflict is indicated in case their minimum distance $d_{min}(s^k_i,s^\lambda_j) \in \mathbb{R}$ is less than $r_{c}$ and $\binom{\mathcal{P}}{2}$ corresponds to the path pairs in $\mathcal{P}$. As a result, the spatio-temporal clearance constraint of the conflicting UAVs is transformed into temporal, since by identifying the segments that are in conflict and handling the time allocation in each segment, the simultaneous presence of the UAVs in each pair is prevented.
\subsubsection{Time of arrival}
 Taking into consideration the time spent in logistics movements required for the UAVs to move between their home locations to the starting points of their manufacturing paths, additional time allocation is added for each task. Specifically, $\tau^{log}_s \in \mathbb{R}$ and $\tau^{log}_e \in \mathbb{R}$ are added before and after the actual starting and end time of printing the specific chunk denoting the motion of the UAV towards the start of the path and the motion towards its next position correspondingly.     
Additionally, as mentioned before, it is assumed that each manufacturing UAV will track its given manufacturing path $P_\kappa$ with a constant speed $V_{ex}$ while extruding material corresponding to smooth deposition and optimal manufacturing results. 
Given that, the time of arrival and departure at  any segment $s^\kappa_i$ of the overall mission can be calculated in a closed form as follows:
\begin{equation}
    t^\kappa_i = S_\kappa + \tau^{log}_s + \sum_{l=0}^i \frac{d^\kappa_l}{V_{ex}}
\end{equation}
denoting the arrival time at the end of the $i$-th segment of  $P_k$.

\subsubsection{Conflict Resolution Constraints}
\begin{figure}
    \centering
    \includegraphics[width=\linewidth]{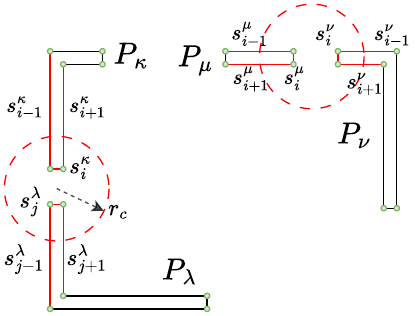}
    \caption{Conflicting segment pairs for minimum clearance of $r_{c}$ m and their arrival times $t^m_i$ for each waypoint.}
    \label{fig:confic_resol_concept}
\end{figure}

As mentioned before, in contrast with \cite{stamatopoulos2024conflict}, the eminent conflicts appearing throughout the execution of the whole mission are not handled by modulating the UAVs velocity (something that would affect the overall printing quality) but by assuming constant speed throughout the whole course and letting the starting time $S_\kappa$ of each task $\mathcal{T}_\kappa$ as free variable. This shifting of the task either backward or forward in time with respect to all the other aforementioned constraints results in the resolution of all the detected conflicts. 
Thus, in the case of a conflicting pair of segments, the convention that the first UAV that will enter the pair will be the first one to leave it, is followed similarly to the one introduced in \cite{stamatopoulos2024conflict}. 
A visual representation of the concept is shown in Fig. \ref{fig:confic_resol_concept} where the two paths $P_\kappa$ and $P_\lambda$ have a conflicting segment pair $(s^\kappa_i,s^\lambda_j)$ (highlighted with red) along with the annotated arrival times at the start of each segment.

The order between the robots is not predefined but is dynamically selected via the solver and for every conflicting pair $(s^k_i,s^\lambda_j)$ and it is denoted using the binary variable $c^{\kappa, \lambda}_{i,j} \in \{0, 1\}$, which is defined as follows:
\begin{equation}
    \label{eq:binaryVar}
    c^{\kappa, \lambda}_{i,j}=
    \begin{cases} 
        1 & ,\text{if }t^\kappa_{i-1} < t^\lambda_{j-1} \\
        0 & ,\text{otherwise}
    \end{cases}
\end{equation}
indicating that if $c^{\kappa, \lambda}_{i,j}=1$, the robot executing path $P_\kappa$ will reach the conflicting pair first and it will occur the other way around in case $c^{\kappa, \lambda}_{i,j}=0$.
Finally, for every pair of segments $ (s^\kappa_i,s^\lambda_j) \in \mathbb{C}$, the aforementioned constraints are imposed via employing the big-M notation \cite{bigM_notation}:
\begin{equation} \label{eq:conflictFreeConstr}
    \mathbb{C}^{\kappa, i}_{\lambda,j}:
    \begin{cases}
        t^k_{i-1} \leq t^\lambda_{j-1} + O(1 - c^{\kappa, \lambda}_{i,j}) \\
        t^k_{i-1} \leq t^\lambda_{j-1}  + O c^{\kappa, \lambda}_{i,j}  \\
        t^k_{i} \leq t^\lambda_{j-1} + O(1 - c^{\kappa, \lambda}_{i,j}) \\
        t^\lambda_{j} \leq t^k_{i-1} + O  c^{\kappa, \lambda}_{i,j}  
    \end{cases}
    ,\forall (s^k_i,s^\lambda_j) \in \mathbb{C} 
\end{equation}
where $O\in \mathbb{R}$ is a sufficiently large number. 

\subsection{Makespan Minimization -  ($\mathbf{P_1}$)}
\label{sec:objectives}
The overall objective of the scheduling algorithm is to accomplish all the printing tasks of set $\mathcal{T}$ and construct 
the given mesh in the least time possible while being with respect to all the aforementioned constraints mentioned in Sections \ref{sec:constraints}  and particularly in making sure no conflict will occur by dynamically deciding the starting time of each task (Section \ref{sec:conflictFreeConstraints}). The overall time span of all the tasks needs to be minimized, translating to 
penalizing the maximum completion time $C_i$ of all the tasks in set $\mathcal{T}$. This is formulated by introducing 
new decision variable $C_{max}$, which denotes the makespan of the whole mission, by adding the following constraint to the overall problem
\begin{equation}
    \label{eq:makespanConstraint}
    S_i + \tau^{log}_s + d_i + \tau^{log}_e \leq C_{max}, \: \forall \mathcal{T}_i \in \mathcal{T}
\end{equation}
which is setting the $C_{max}$ as the upper bound of all the completion times of tasks. Finally, the term $C_{max}$ is added to the objective function as the main objective term $J_{ms} = C_{max}$. 

The optimization problem incorporating the aforementioned and the conflict-free safety constraints (analyzed in Section \ref{sec:conflictFreeConstraints}) along with the makespan minimization objective constitutes the makespan minimization scheduling problem denoted as $\mathbf{P}_1$ and is sufficient to generate a mission plan with respect to the construction mission requirements.

\subsection{Task Importance Prioritization -  ($\mathbf{P_2}$)}
Trying to prioritize the tasks that have the most dependencies and alleviate the scheduling from possible bottlenecks while also imposing a faster convergence to the overall optimization problem, some heuristics are added in the overall cost function with a smaller priority. Specifically,
despite the fact that the makespan of the overall mission is penalized following the approach of a min-max problem, the sum of all the ending times $E_i$ of tasks $\mathcal{T}_i$ is penalized, but each one is scaled using an importance factor $\alpha_i \in \mathbb{R}$. The importance factor $\alpha_i$ signifies the contribution of a particular task $\mathcal{T}_i$ to the overall construction of the mesh $\mathcal{M}$. This is translated into the in-degree $d^-(C_i)$ of a node $C_i$ of dependency graph $G^d$, which notates the number of incoming edges/ancestors to a node of the graph and is defined as follows:
\begin{equation}
     d^-(C_i) = \sum_{u \in \mathcal{V}^d} \delta{(u, C_i)}
\end{equation}

where $\mathcal{V}^d$ is the set of all nodes of dependency graph $G^d$ and 
\begin{equation}
    \delta{(u, v)} =
    \begin{cases} 
        1 & \text{if } (u, v) \in \mathcal{E}^d, \\
        0 & \text{otherwise.}
    \end{cases}
\end{equation}
where $\mathcal{E}^d$ denotes the set of edges of the directional graph $G^d$.

Following a similar approach, the second-order in-degree is defined for a vertex $C_i$ of graph $G^d$ as follows:
\begin{equation}
      d_\mathbb{2}^-(C_i) = d^-(C_i) + \beta \sum_{u \in \mathcal{N}^{-}_{C_i}} d^-(C_i)
\end{equation}
where $\mathcal{N}^{-}_{C_i}$ denotes the set of all the ancestor nodes of node $C_i$ and $\beta \in \mathbb{R}$ is a tunable scaling factor.
So, the importance factor $\alpha_i = d_\mathbb{2}^-(C_i)$ and the importance score is calculated as follows:
\begin{equation}
    J_{im} = \sum_{i=0}^{N-1}  a_i (S_i +\tau^{log}_s + d_i + \tau^{log}_e)
\end{equation}
The importance score $J_{im}$ is scaled with $W_{im}$ and added to the overall objective function of the problem $J=W_{ms}J_{ms} + W_{im}J_{im} $, leading to the problem variation $\mathbf{P}_2$.
\begin{figure}
    \centering
    \includegraphics[width=0.95\linewidth]{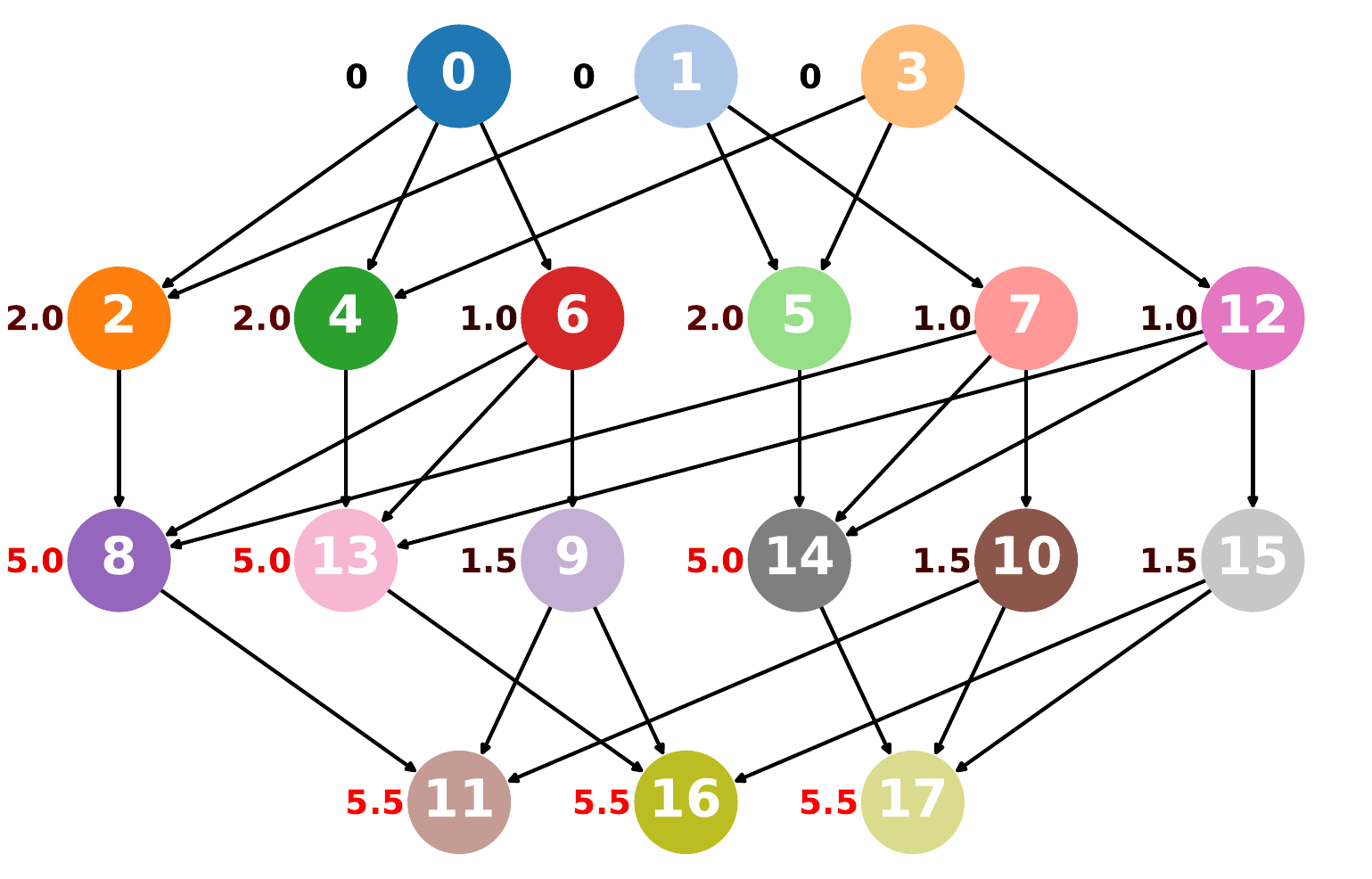}
    \caption{Dependency graph along with annotated importance $\alpha_i$ for each node.}
    \label{fig:importanceCosts}
\end{figure}

\subsection{Agent Utilization -  ($\mathbf{P_3}$)} \label{sec:agentUtil}
After analyzing the scheduling results, it has been observed that the makespan does not continue to decrease beyond a certain point, despite an increase in the number of available UAVs. This phenomenon arises primarily due to task dependencies and, more critically, conflicts between tasks. Although UAVs and tasks are available at any given moment, deploying certain tasks may violate the minimum clearance requirements between UAVs. Consequently, these tasks are held off until the completion of higher-priority tasks. As a result, multiple UAVs often remain landed and idle while awaiting the completion of preceding and conflicting tasks.

Inefficient deployment of UAVs is undesirable, as it incurs unnecessary costs and maintenance burdens. To address this issue, an additional objective has been introduced into the problem formulation: determining the optimal number of UAVs required for a given mission. This approach aims to eliminate the deployment of excess agents while balancing the trade-off between makespan and UAV utilization.

In order to facilitate this, a cost on using each robot $R_k$ is included. It is assumed that a robot is utilized in the mission if it has been assigned at least one task $\mathcal{T}_i$ throughout the whole duration of the mission. The utilization of the robot $R_k$ is notated using the binary variable $u_k \in \{0, 1\}$ which is defined as follows:
\begin{equation}
    u_k = 
    \begin{cases} 
        1 & \text{if } \exists i : x_{i,k} = 1, \\
        0 & \text{otherwise}.
    \end{cases}
\end{equation}
where $x_{i,k}$ indicates whether task $\mathcal{T}_i$ is assigned to robot $R_k$.

This switching case is mathematically formulated in the optimization problem using the big-M notation by adding the following two constraints.

\begin{equation}
    \label{eq:agentUtilConstr}
    \begin{cases}
      \sum_{i=0}^{N-1} x_{i,k} \leq O u_k \\
      \sum_{i=0}^{N-1} x_{i,k} \geq u_k \\
    \end{cases}
    ,\forall u_k \in \mathcal{U}
\end{equation}
where $\mathcal{U} = \{u_i| i = 0, 1, ..., M\}$ is the set containing all the binary utilization variables corresponding to each robot $R_k$.

Finally, the agent utilization cost is formulated as follows:
\begin{equation}
    J_{ut} = \sum_{k=0}^{M-1} W_{cost} u_k
\end{equation}

 where $W_{cost} \in \mathbb{R}$ is a tunable cost associated with the use of an agent in the mission. 
 With the addition of this final objective, the problem variation aligning with all the contributions of this paper is completed and is referred to as $\mathbf{P}_3$.

\subsection{Problem Overview}
The aforementioned costs are fused into a common one $J$ scaled by custom gains $G_{ms}, G_{pr}, G_{ut} \in \mathbb{R}$ as follows:
\begin{equation}
    J = G_{ms} J_{ms} + G_{im} J_{im} + G_{ut} J_{ut}
\end{equation}
This cost is fed to the optimization problem in order to be minimized with respect to the constraints.
An overview of the whole problem formulation is given as follows:

   

\begin{equation}
\begin{aligned}
\argmin_{x_{j,i},\, y^{\kappa,i}_{\lambda,j},\, C_{\max}} \quad 
    & G_{ms} J_{ms} + G_{im} J_{im} + G_{ut} J_{ut} \\
\text{subject to:} \\
\forall\, \mathcal{T}_i \in \mathcal{T}: \quad 
    & \text{Eqs.~(\ref{eq:startTime}), (\ref{eq:assigment}), (\ref{eq:makespanConstraint})} \\
\forall\, \mathcal{T}_i \neq \mathcal{T}_j \in \mathcal{T},\; R_k \in \mathcal{R}: \quad
    & \text{Eqs.~(\ref{eq:ordering1}), (\ref{eq:ordering2})} \\
\forall\, (\mathcal{T}_i,\mathcal{T}_j) \in \mathcal{P}_{pr}: \quad
    & \text{Eq.~(\ref{eq:precedence})} \\
\forall\, R_k \in \mathcal{R}: \quad
    & \text{Eqs.~(\ref{eq:materialBudget}), (\ref{eq:timeBudget})} \\
\forall\, (s^k_i, s^\lambda_j) \in \mathbb{C}: \quad
    & \text{Eq.~(\ref{eq:conflictFreeConstr})} \\
\forall\, u_k \in \mathcal{U}: \quad
    & \text{Eq.~(\ref{eq:agentUtilConstr})}
\end{aligned}
\end{equation}

The problem is a Mixed Integer Programming (MIP) and its formulation along with its solution is carried out using the Gurobi Optimization solver \cite{gurobi}.

\subsection{Mission Execution}
\subsubsection{Mission Control}

Once the optimal schedule is generated based on the mission requirements, chunking, and UAV configuration, the schedule is sent to the mission planner. The planner converts it into a time-based event log, specifying the timing for each task $\mathcal{T}_i$ and its assigned robot $R_k$. At the mission's start, mission control transmits a synchronization signal to all agents, instructing them to set their internal timers to $t = 0$. As time progresses, tasks are triggered according to the schedule. When the time for a new task begins, mission control assigns the corresponding task $\mathcal{T}_i$ to the designated agent $R_k$ based on the plan.

\subsubsection{Printing UAV}

Whenever a UAV $R_k$ is assigned a new task, it moves to the starting point of the manufacturing path. 
\textcolor{blue}{
Both the logistics start and end times, $\tau^{\text{log}}_s$ and $\tau^{\text{log}}_e$, are defined as uniform constants across all UAVs. 
%
Under this assumption, it is possible that one UAV may arrive at its designated start position earlier than scheduled. In such cases, the UAV hovers in place until synchronization with the global schedule is restored. This hovering period represents only a small fraction of the total logistics time $\tau^{\text{log}}_s$ and is negligible compared to the overall task durations and the total mission makespan.
}
%
During the printing process, the UAV follows the manufacturing path for the task $\mathcal{T}_i$ as generated by the slicer, depositing material at a constant speed $V_{ex}$. Once printing is complete, the UAV returns to its designated landing position and remains idle until assigned another task.

While traveling between the home positions and the printing workspace, UAVs fly at a cruising altitude $h_{cr}$ to prevent the downdraft from their propellers from interfering with previously printed chunks. Collision avoidance during logistics movements is decentralized: UAVs share their planned trajectories, similar to the approach in \cite{bjornCollisionAvoidance}. Each UAV treats the trajectories of others as obstacles in space-time and replans its own trajectory in case a potential collision is detected.

\section{Results}\label{sec:results}

The evaluation of the framework is carried out in various meshes and the main focus is on a rectangle one with dimensions of $2$ m. width, $2$ m. length and $0.5$ m. height. It is decomposed into $18$ chunks shown in Fig. \ref{fig:chunksFeatures}.a and the dependencies between them are calculated and placed in the dependency graph $G^d$ shown in Fig. \ref{fig:chunksFeatures}.b. The parameters used throughout the formulation of the optimization problem for the rectangle scheduling, along with the ones used for the construction mission simulation, are listed in Table \ref{tab:parametersValues}.

\begin{table}[h]
\centering
\resizebox{0.8\linewidth}{!}{ 
    \begin{tabular}{ccccc}
    \hline
    \multicolumn{1}{c}{Parameter} & \multicolumn{1}{c}{Value} & \multicolumn{1}{c}{} & \multicolumn{1}{c}{Parameter} & \multicolumn{1}{c}{Value} 
    \\ 
    \hline    
    $\tau^{log}_s$ & $15$ sec. &   & $\beta$   & $0.5$    \\
    $\tau^{log}_e$ & $15$ sec. &   & $G_{ms}$  & $1$    \\
    $V_{ex}$    & $0.1$ m/s &   & $G_{im}$  & $0.07$  \\
    $r_c$       & $1$ m.    &   & $G_{ut}$  & $100$  \\
    $\alpha$    & $1$       &   & $$& $$     \\\hline
    \end{tabular}
}
\caption{Parameter values for the formulation of the optimization scheduling problem.}
\label{tab:parametersValues}
\end{table}

\begin{figure*}[b]
    \centering
    \includegraphics[width=\linewidth]{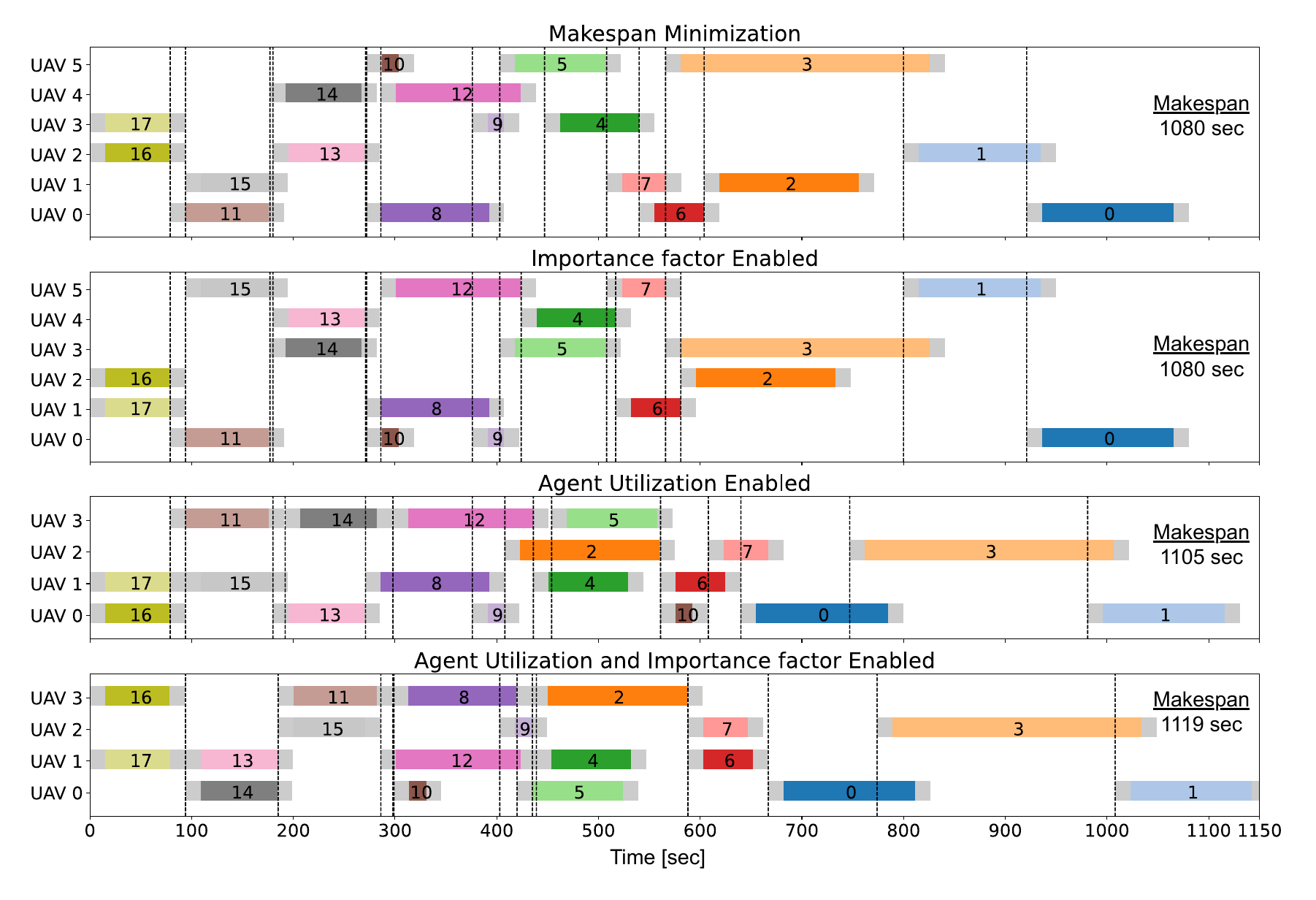}
    \caption{{Rectangular mesh scheduling results for different formulations of the objective function.}}
    \label{fig:allSchedulingResults}
\end{figure*}

The optimal scheduling results are presented as resource utilization charts similar to the ones shown in Fig. \ref{fig:allSchedulingResults}, where the $y$-axis represents the available UAVs, and the $x$-axis corresponds to time. Each task is represented by color-coded blocks, labeled with the respective task index, indicating the time periods during which each UAV is assigned a printing task and the time assigned to logistics is annotated with gray color.
\textcolor{blue}{Aligned with previous iterations of aerial 3D printing such as \cite{nature_aerial_AM, stamatopoulosICUAS} along with aerial masonry construction \cite{ETHAerialMasonry, metaheuristicAerialMasonry, stamatopoulos2025Masonry}, an available team of $6$ UAVs $\mathcal{R}= \{R_0, R_1, ..., R_5 \}$ is used for all cases of the rectangular mesh and the various problem formulations are evaluated and analyzed}
%
%
At the end, a case study is described and demonstrated as a complete construction mission in a simulated Gazebo environment with autonomous aerial robots.


\begin{table*}[t]
\centering
\resizebox{0.9\linewidth}{!}{ %
\begin{tabular}{ccccccc}
\hline
    \begin{tabular}[c]{@{}c@{}}Importance \\ Cost\end{tabular}              &
    \begin{tabular}[c]{@{}c@{}}Agent Utilization \\ Cost\end{tabular}       & 
    \begin{tabular}[c]{@{}c@{}}Available \\ UAVs \end{tabular}          &
    \begin{tabular}[c]{@{}c@{}}Total Number of \\ UAVs Used\end{tabular}    & 
    \begin{tabular}[c]{@{}c@{}}Material Usage \\ $[$L$]$\end{tabular}    & 
    \begin{tabular}[c]{@{}c@{}}Computation \\ Time [sec]\end{tabular}       & 
    \begin{tabular}[c]{@{}c@{}}Overall Solution \\ Makespan [sec]\end{tabular}
    
    \\[8pt] \hline
No   & No    & 6 & 6 & 
\begin{tabular}[c]{@{}c@{}}[88, 61, 63, 37, 54, 74]\end{tabular} 
& 119.04 & 1080.50 \\[6pt]
Yes  & No    & 6 & 6 & 
\begin{tabular}[c]{@{}c@{}}[56, 57, 47, 91, 36, 91]\end{tabular} 
& 99.76  & 1080.50 \\[6pt]
No   & Yes   & 6 & 4 & 
\begin{tabular}[c]{@{}c@{}}[96, 93, 95, 93]\end{tabular}       
& 244.49 & 1130.58 \\[6pt]
Yes  & Yes   & 6 & 4 & 
\begin{tabular}[c]{@{}c@{}}[92, 99, 89, 98]\end{tabular}       
& 216.88 & 1157.58 \\

\hline
\end{tabular}
}
\caption{{Makespan for the rectangle mission and computational time required for different cost objective formulations}}
\label{tab:compTimes}
\end{table*}

\subsection{Effect of Task Importance Factor}
To emphasize the impact of task importance on the overall solution of the optimization problem, the case of the rectangle illustrated in Fig.~\ref{fig:chunksFeatures}.a is evaluated both with and without the inclusion of the task importance cost in the objective function. The importance cost for each task, $\mathcal{T}_i$, associated with the chunks of the rectangle is depicted in Fig.~\ref{fig:importanceCosts}. These costs are shown to the left of each node and are visually represented using a color gradient ranging from black (indicating zero importance) to red (indicating higher values).

Figure~\ref{fig:allSchedulingResults} presents the mission scheduling solutions, where each pair of plots corresponds to one run without and one with task importance penalization. The first two plots show results without considering agent utilization cost, while the last two rows include it as part of the objective function (corresponding to problem $\mathbf{P}_3$).
{
For the case without agent utilization, it is noted that the assignment of the tasks to the available robots has changed; however, the sequence between them has remained the same despite being assigned to different UAVs. Additionally, it is noted that the starting time of some tasks has been placed earlier compared to the solution without the importance factor. Specifically, the tasks $\mathcal{T}_2$, $\mathcal{T}_4$ and $\mathcal{T}_6$ have been calculated to start earlier compared to the base solution by $23$ \si{\sec}. The overall solution makespan of the planned mission for both cases (with and without importance factor included), along with the computational time required to calculate the optimal plan, are shown in the first and second rows of the Table \ref{tab:compTimes}. It is noticed that although the makespan remains the same, the main advantage of including the task importance is the faster convergence and thus the reduction in time of the whole computation of the optimal solution from the solver. Evidently, with the inclusion of the importance cost, the computation time is decreased by $16\%$ (from $119.04$ \si{\sec} to $99.76$ \si{\sec}) in the case of adding only the importance cost (problem $\mathbf{P}_2$).}

{
In the last two plots, where the case of agent utilization is involved (problem $\mathbf{P}_3$), it is noticed that tasks of high importance are prioritized towards the start of the mission, while others of least importance are placed towards the end. Task $\mathcal{T}_{13}$ with an importance factor of $\alpha_{13} = 5.0$ is executed $86$ \si{\sec} earlier similar to task $\mathcal{T}_{10}$, of importance equal to $\alpha_{10} = 1.5$, which is also placed $262$ \si{\sec} earlier compared to the original solution.
Focusing on the mission makespan and the computation time, it is noticed that there has been a $2.38\%$ increase in the makespan, but at the same time, the computation time has been decreased by $12\%$. Although the makespan is slightly increased, the main advantage of including the task importance is the faster convergence and thus the reduction in time of the whole computation of the optimal solution from the solver. 
Overall, the problem formulation has been formulated to accommodate and adapt to various application scenarios. In cases where the primary objective is to minimize mission duration without accounting for computational cost, the importance factor can be omitted from the objective function. This design choice ensures that the model remains adaptable to varying operational requirements.

\subsection{Dynamic sized team of UAVs}

Aiming to evaluate the effectiveness of the framework on selecting the optimal number of agents for the given tasks $\mathcal{T}$ and the given UAVs configuration, the problem $\mathbf{P}_3$ of the scheduling optimization is solved by providing it with six available UAVs ($M=6$) {each one of them equipped with $10$ L. of material, resulting in $100$ L. of actually expanded, fully cured mesh)} and a robot utilization cost $G_{ut}=100$. As shown in Fig. \ref{fig:allSchedulingResults}, the makespan of the mission remains almost the same, while only four UAVs are now used instead of the six originally available. The amount of time the UAVs remain idle is significantly reduced.
%
%
To demonstrate further the effectiveness of this capability, an investigation on associating the number of the total UAVs $M$ with the total makespan of the mission $C_{\text{max}}$ is carried out. Specifically, using the problem formulation $\mathbf{P}_1$, the optimal scheduling plan was generated for the mission using varying numbers of robots, from a single robot ($M=1$) to six ($M=8$), while considering that both the material they are carrying along with their battery are sufficient. The resulting mission makespans are shown in Fig. \ref{fig:makespanAgentsScaling}. While the mission duration decreases as more UAVs are added, this reduction plateaus beyond a certain point. This is due to conflicts arising from multiple UAVs operating simultaneously in a confined space, leading to increased wait times and under-utilization of available resources. This behavior is also evident in Fig. \ref{fig:allSchedulingResults}, where UAVs $R_4$ and $R_3$ are only assigned to two and three tasks each. Additionally, no more than three UAVs are actively working at the same time during the majority of the mission. These findings suggest that fewer UAVs could be deployed for this specific mission without significantly impacting its overall performance in terms of time.
\begin{figure}[b]
    \centering
    \includegraphics[width=\linewidth]{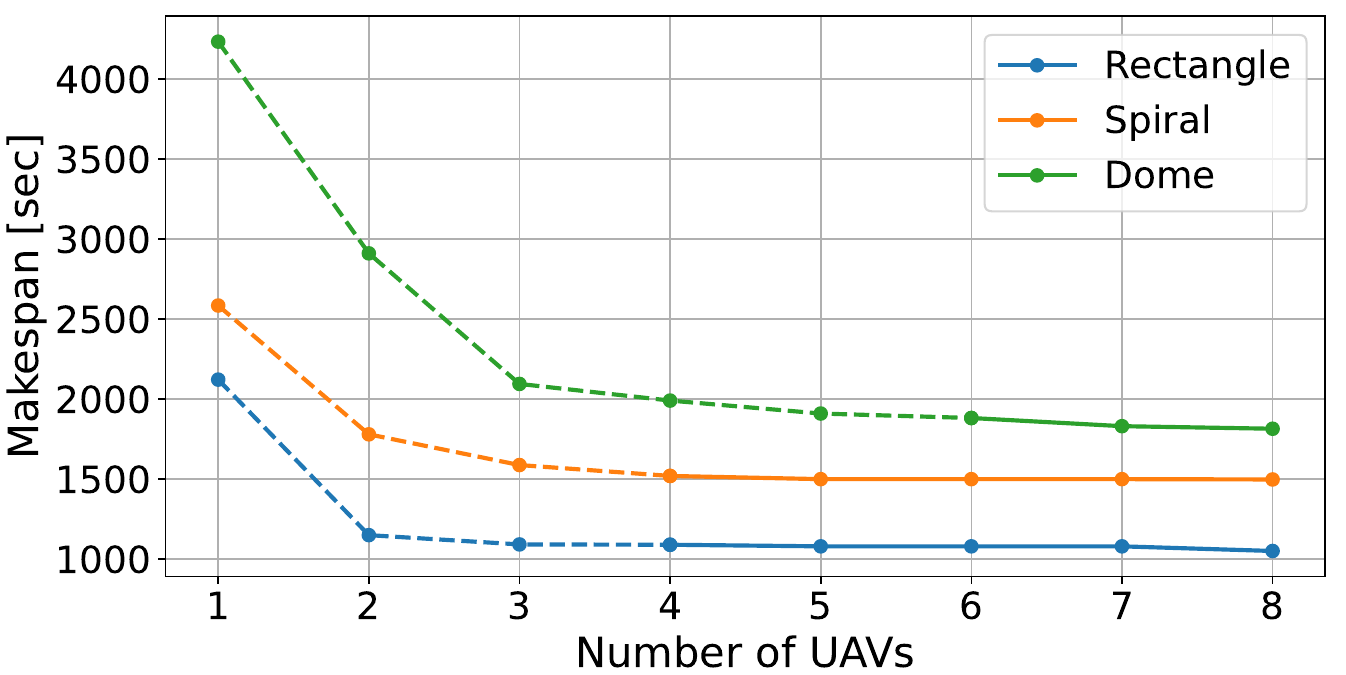 }
    \caption{
    Mission makespan for rectangular, dome, and quatrefoil mesh configurations with varying numbers of UAVs. Plot lines corresponding to configurations that violate the material constraint are rendered as dotted.
    }
    \label{fig:makespanAgentsScaling}
\end{figure}

{
The final calculated mission makespans for the two versions of problem $\mathbf{P}_3$ (with and without the importance factor), along with the computational times required to solve them, are presented in the last two rows of Table \ref{tab:compTimes}. Compared to problem $\mathbf{P}_1$, the makespan is increased by $4.4\%$ in the case without and $7.1\%$ in the case with the importance cost included.
This slight increase is considered to be negligible given the fact that it is achieved with two UAVs less and at the same time maximizing the utilization of the remaining four that are deployed.} 
Although the computational time increases by $105.0\%$ and $81.5\%$, respectively is significant, the algorithm remains efficient compared to solving problem $\mathbf{P}_1$ independently for different instances, where the number of UAVs $M$ ranges from $1$ to $6$.
This search process would require at least $3$ different computations of the problem in order to dynamically identify where the slope of the plot shown in Fig. \ref{fig:makespanAgentsScaling} is no longer being decreased. The latter approach requires a total computation time of $607$ seconds, which is notably higher.
It is highlighted that the algorithm dynamically identifies the optimal number of UAVs for the mission. This is evident in Fig. \ref{fig:makespanAgentsScaling}, where the makespan shows no significant decrease once $M$ exceeds $4$.
It must be noted that, although it may still seem advantageous to deploy only three UAVs, reducing the fleet below four leads to violations of the material budget. This is visually indicated in the plot using a dotted line to represent configurations that exceed the material constraints.
%


\textcolor{blue}{
The trade-off between agent utilization and makespan reduction is governed by the utilization factor and the gain parameters $G_{\text{ms}}$ and $G_{\text{ut}}$. Adjusting these weights alters the number of UAVs deployed, thereby influencing the overall mission duration. Increasing the value of $G_{\text{ut}}$ places a stronger penalty on activating additional UAVs, which leads the optimizer to employ fewer agents. Consequently, the total makespan increases since fewer UAVs share the workload. Conversely, reducing $G_{\text{ut}}$ promotes the use of more agents, resulting in faster mission completion at the cost of higher resource utilization. The framework aims to identify the equilibrium point where further reductions in makespan no longer justify deploying additional UAVs, that is, the point where the makespan curve levels off, signifying a configuration that jointly minimizes both fleet size and mission duration.
In practical applications, emphasizing $G_{\text{ms}}$ is appropriate when minimizing mission duration is critical (e.g., in time-sensitive operations), while increasing $G_{\text{ut}}$ encourages more balanced workload distribution and improved resource efficiency. Our analysis suggests that maintaining a ratio of $G_{\text{ms}} J_{\text{ms}}$ to $G_{\text{ut}} J_{\text{ut}}$ between 2:1 and 3:1 is effective, allowing the optimization to prioritize makespan reduction while still accounting for fleet size minimization.
If the task-importance term is included, its influence should remain small relative to the makespan term to prevent it from dominating the objective function. This ensures that it serves primarily to aid convergence by giving slight priority to more critical tasks, without disturbing the fundamental balance between mission duration and fleet utilization.
}

The trade-off between the number of agents and makespan minimization, while simultaneously satisfying budget constraints, is addressed within the optimization process in a single step.
This capability is particularly advantageous, as the cost of introducing additional UAVs can be substantial. Selecting the optimal number of UAVs required for the task not only minimizes construction costs but also reduces the complexity associated with managing the equipment, thereby improving the overall feasibility and efficiency of the operation.

\subsection{Evaluation on Different Mesh Sizes}

\begin{table*}[b]
\centering
\resizebox{\linewidth}{!}{ 
\begin{tabular}{rccccccc} 
\hline
\multicolumn{1}{c}{Case} &
    \begin{tabular}[c]{@{}c@{}}Total Number of \\ Chunks\end{tabular}    & 
    \begin{tabular}[c]{@{}c@{}}Total Number of \\ UAVs Used\end{tabular}    & 
    \begin{tabular}[c]{@{}c@{}}Number of \\ Dependencies\end{tabular}       &
    \begin{tabular}[c]{@{}c@{}}Number of \\ Conflicts\end{tabular}          &
    \makecell{Conflict Density \\ {[conflicts/$m^2$]}} &
    \begin{tabular}[c]{@{}c@{}}Computation \\ Time [sec]\end{tabular}       & 
    \begin{tabular}[c]{@{}c@{}}Overall Solution \\ Makespan [sec]\end{tabular} 
    \\ \hline
(I) 2x2 m.             & 18  & 6 & 30  & 27107  & 6776  & 156.6    & 1092 \\
{(II) 2x2 m.} & {30}  & {6} & {50}  & {35026} & {8756}  & {1330.85}  & {1105} \\
(III) 3x3 m.           & 55  & 6 & 70  & 44664  & 4962    & 846.4    & 1650 \\
(IV) 4x4 m.            & 60  & 6 & 109 & 51301  & 3206    & 2138.6   & 1830 \\
\hline
\end{tabular}
}
\caption{Evaluation of the optimal scheduling for different sizes and chunking of meshes.}
\label{tab:compTimesDiffSizes}
\vspace{-5mm}
\end{table*}

To evaluate and analyze how the scheduling algorithm responds and adapts to meshes of different sizes, the construction plan is computed for the same mesh scaled into four different cases ($1-4$): rectangles with sides measuring $2$, $3$, and $4$ meters, respectively. The meshes are processed through the chunker module \cite{stamatopoulosChunkingJINT} and subsequently fed into the scheduling module. The results of the evaluation are summarized in Table \ref{tab:compTimesDiffSizes}.
As anticipated, the number of chunks $|\mathcal{C}|$ increases with the size of the mesh, along with the number of dependencies $|\mathcal{G}_{d}|$ and conflicts $|\mathbb{C}|$. However, it is noteworthy that the increase in conflicts ($164\%$ and $189\%$ for cases $3$ and $4$ compared to case $1$) is lower than the corresponding increase in dependencies ($233\%$ and $363\%$, respectively). This can be attributed to the fact that, despite the scaling of the mesh $\mathcal{M}$, conflicts remain primarily localized between neighboring chunks due to the safety clearance $r_c$. Consequently, the addition of new chunks located farther apart does not necessarily result in a proportional increase in conflicts. The decomposition performed by the chunker \cite{stamatopoulosChunkingJINT} has a significant impact on the number of dependencies and conflicts among the tasks. 
{%
This is evident in a separate evaluation in which the original rectangular mesh was further partitioned into thirty chunks. As anticipated, the planning complexity increased: the number of dependencies rose to fifty, and the number of conflict pairs reached 35026. However, this finer partitioning did not grant additional flexibility, since conflict resolution is performed at the path‐segment level rather than at the chunk level. As a result, the makespan (1105.5\,\si{\second}) remained comparable to that (1080.5\,\si{\second}) of the original decomposition, and was in fact slightly longer due to the greater logistical overhead between tasks. Moreover, computational time increased significantly ($10.45$ times) with the added complexity, yet without producing any meaningful reduction in the makespan.
Despite the aforementioned analysis, enhancements in the chunking module, along with finding the optimal number of chunks for a given mission, are beyond the scope of the current paper. 
}

\textcolor{blue}{In general, Table \ref{tab:compTimesDiffSizes} further crystallizes the comparative analysis of scheduling performance across different workspace sizes, highlighting how spatial constraints influence both agent coordination and overall mission complexity. While larger meshes exhibit more tasks and dependencies, they also allow greater maneuverability and lower conflict densities, reflecting improved spatial separation between agents. In contrast, smaller workspaces exhibit a much higher concentration of conflicts relative to their area, indicating increased interaction among agents and tighter safety margins. These findings reinforce that compact construction footprints, where agents operate within overlapping working envelopes, constitute the most demanding scenarios. It is under such tight-space conditions that the advantages of the proposed scheduling and coordination framework become evident, as the system must continuously balance safety, dependency, and efficiency constraints in a highly constrained environment.
}

\begin{figure*}
    \centering 
    \includegraphics[width=0.9\linewidth]{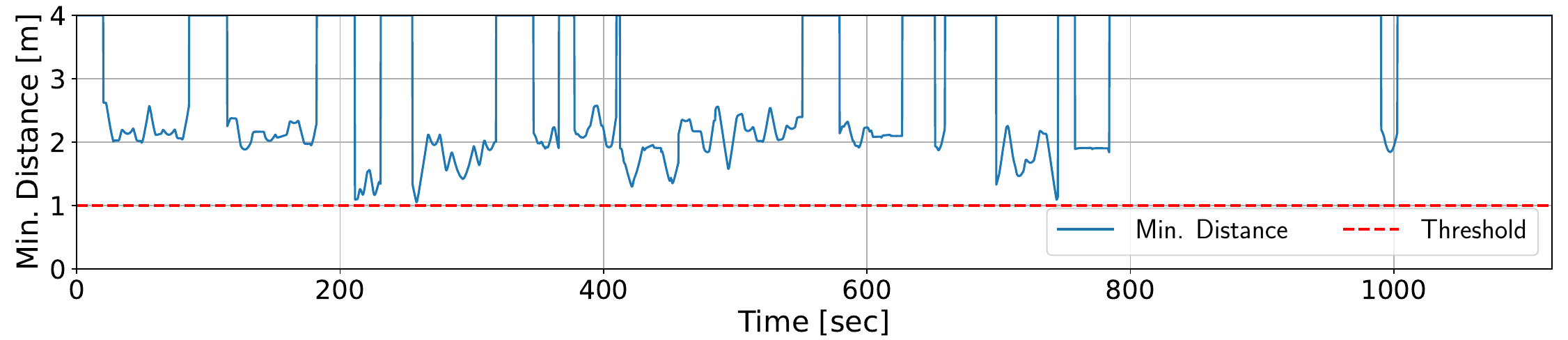}
    \caption{Minimum Distance (blue) between UAVs during the execution of the whole mission.}
    \label{fig:minUAVsDist}
    \vspace{-5mm}
\end{figure*}

\subsection{Case Study - Construction Mission Simulation}
To evaluate the deployment and integration of the optimal multi-agent aerial additive manufacturing framework in a complete construction mission, a simulation is conducted using a specific case study scenario. The framework is tested within a Gazebo simulation environment, where multiple UAVs, modeled after designs in \cite{carma,stamatopoulosICUAS}, are equipped with a bottom-mounted extruder suspended beneath them. The tip of each extruder is precisely tracked in 3D space, and material deposition is visualized by placing small spherical markers directly below the extruder tip whenever a deposition action is triggered by the corresponding UAV.

The construction scenario of the case mentioned in Section \ref{sec:agentUtil} is investigated. In the specific one, the rectangular mesh of Fig. \ref{fig:chunksFeatures}.a is fed to the scheduling module and six UAVs constitute the 
available resources for the mission. Specifically, following previous iterations of similar concepts \cite{nature_aerial_AM, kovac}, it is assumed that the material carried by each UAV is a light-weight expandable polyurethane foam due to its main advantage of expansion (assumed an expansion factor of $10$ \cite{foamExpansion}) right after it is sprayed and has come into contact with air. The carried volume of each UAV is set to {$10$} L while the flight time $t^b_k$ of each UAV is set to $15$ min. The home positions of the UAVS are placed in the corners of a square with a size of $6$ m. while the cruising altitude for the mission was selected to be equal to $h_{cr} = 1.5$ m. The duration for carrying out the logistics right before and after the execution of each task is selected as $\tau^{log}_s=\tau^{log}_e=15$ sec after experimentation and measuring the maximum duration of the motion from the home position of a UAV to the starting point of its assigned task. The minimum clearance $r_c$ between the UAVs while printing is selected to be equal to $1$ m.

As discussed in the previous section and shown in Fig. \ref{fig:makespanAgentsScaling}, the optimal plan generated by the scheduling module solving the problem $\mathbf{P}_3$ utilizes only four of the six available robots, reducing the resource cost by $33\%$. This plan is then passed to the mission planner, initiating the mission and ensuring all UAVs remain in direct communication and synchronized. The minimum clearance $r_c$ between UAVs is maintained throughout the mission, as demonstrated in Fig. \ref{fig:minUAVsDist}, highlighting the framework's capability to allocate multiple parallel printing tasks while ensuring safety.
It is observed that, while tasks $\mathcal{T}_1$ and $\mathcal{T}_3$ can theoretically be printed simultaneously from a dependency perspective, they are not scheduled to start together due to potential conflicts between them. Specifically, task $\mathcal{T}_1$ is scheduled to start at \( t = 1008 \, \text{sec} \), after the conflicting segments of task $\mathcal{T}_3$ have already been printed, ensuring that any potential safety risks are eliminated. {A similar case is observed between tasks $\mathcal{T}_0$ and $\mathcal{T}_3$ where although they can be printed in parallel as well, concerning their structural dependencies, the starting time of task $\mathcal{T}_3$ is delayed at $  t = 774 \, \text{sec}$ to resolve the conflicts with $\mathcal{T}_0$.}
\textcolor{blue}{The conflict among tasks $\mathcal{T}_0$, $\mathcal{T}_1$, and $\mathcal{T}_3$ is consistently evident across all cases shown in Fig.~\ref{fig:allSchedulingResults}. It can be observed that their conflict-free simultaneous execution is infeasible; hence, the algorithm schedules them sequentially, permitting only minimal temporal overlap. This behavior remains consistent regardless of the number of UAVs or the total payload, as illustrated across all scenarios where these tasks are invariably assigned in sequence with slight overlaps at their start and end times, showcasing that this constraint is associated with the shape's and the chunks' geometry.
Notably, this scheduling outcome is an inherent result of the algorithm’s conflict-resolution mechanism rather than any manual or static adjustment, reflecting constraints imposed by the geometry of the shape and the corresponding chunks.
}
The total volume of material used by each UAV for the mission is as follows: $[9.2, 9.9, 8.9, 9.8]]$~L, remaining within the initial material budget $\mathcal{V}^R$. Similarly, the total flight time for each UAV is $[9.66, 8.96, 8.25, 8.48]$~minutes, remaining below the flight time limit.
A video of the whole simulated mission, including a dynamic visualization of both the dependency and conflict graphs throughout the mission, is available in the following link: \href{https://youtu.be/b4jwhkNPTyQ}{https://youtu.be/b4jwhkNPTyQ}.

\begin{figure*}[t]
    \centering
    \includegraphics[width=0.9\linewidth]{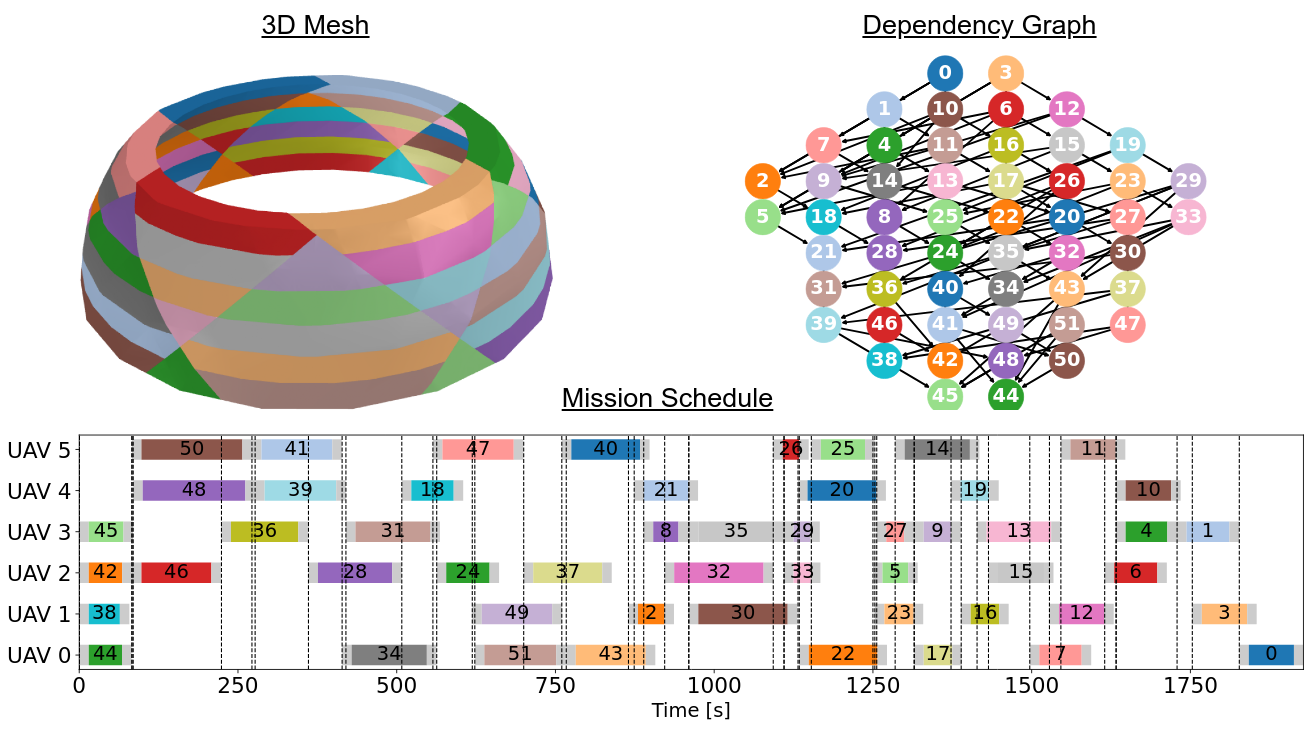}
    \vspace{-2mm}
    \caption{{Hemispherical Dome mesh of \SI{3}{\metre} diameter and a height of \SI{1}{\metre} decomposed in $52$ chunks (top left) along with its associated dependency graph $G^d$ (top right). The optimal schedule for the mesh and a fleet of $10$ UAVs is shown at the bottom. }}
    \label{fig:domeFig}
\end{figure*}

\begin{figure*}[b]
    \centering
    \includegraphics[width=0.85 \linewidth]{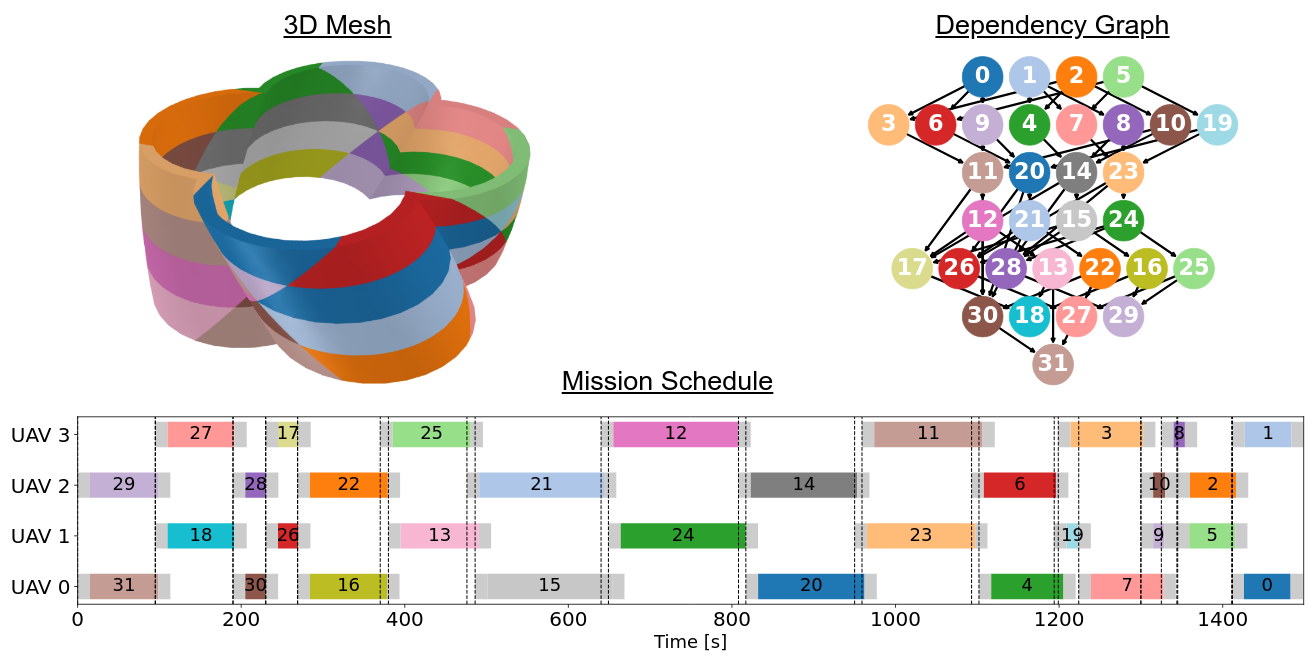}
    \vspace{-2mm}
    \caption{{Quatrefoil (four-lobed) shell mesh of \SI{2}{\metre} side and a height of \SI{0.8}{\metre} decomposed in $32$ chunks (top left) along with its associated dependency graph $G^d$ (top right) and the optimal schedule for the mesh with a team of $8$ UAVs (bottom). }}
    \label{fig:futureFig}
\end{figure*}
\subsection{{Evaluation on Different Meshes}}
{
Towards the further evaluation of the proposed framework for different geometries and meshes, two complex meshes are fed as input. 
Specifically, a hemispherical dome of $3$ m. diameter and a height of  $0.75$ m. is evaluated. The dome is decomposed into $53$ chunks in total, which are shown in a color-coded format in Fig. \ref{fig:domeFig} in a 3D format. Additionally, the associated calculated dependencies between the chunks are placed in the dependency graph and visualized. Finally, the optimal mission schedule for the given mission and for {$8$} UAVs available in total, carrying $20$ L. of material each, is calculated and shown at the bottom of Fig. \ref{fig:domeFig}. The optimal schedule employs $6$ UAVs from the given fleet, resulting in a total mission makespan of \SI{1928}{\second}.
}

{
In addition, an extruded quatrefoil (four-lobed) shell with uniform wall thickness, a height of \SI{0.8}{\metre}, and a side length of \SI{2.0}{\metre} is considered. The shell is partitioned into $32$ chunks, and its corresponding dependency graph, $G^{d}$, with color-coded chunks, is depicted in Fig.~\ref{fig:futureFig}. Upon calculation of each chunk’s manufacturing path and generation of both the dependency and conflict graphs, the configuration, together with eight UAVs, each possessing a \SI{20}{\litre} capacity, is fed to the optimizer. The resulting optimal schedule, also shown in Fig.~\ref{fig:futureFig}, identifies four UAVs as the optimal fleet size, yielding a makespan of \SI{1499}{\second}.
}

\subsection{\textcolor{blue}{Limitations}}
\textcolor{blue}{Despite the effectiveness of the proposed framework, there are still some hurdles that limit its full potential.}
\textcolor{blue}{It is noted that the estimation of the flight time employs a simplified model that establishes a linear relationship between each UAV’s battery status and its estimated flight duration for a given mission. In practice, the actual flight time exhibits nonlinear dependence on factors such as the instantaneous payload weight. However, incorporating such nonlinear modeling would considerably increase the complexity of the overall formulation. Therefore, a simplified linear approximation is adopted, which remains valid in scenarios where the payload variations among UAVs are relatively small.}

\textcolor{blue}{
In the proposed formulation, the logistical travel times $\tau^{\text{log}}_s$ and $\tau^{\text{log}}_e$ for each UAV are assumed constant during motion between tasks and their designated landing point. In practice, however, this assumption does not strictly hold, as each UAV starts from a different initial position and the distances between consecutive tasks vary with the assigned sequence. Incorporating these varying travel times into the optimization would significantly increase computational complexity. Furthermore, since trajectory sharing is employed to prevent inter-UAV collisions during transit, the actual travel time becomes non-deterministic. These effects are therefore neglected, as the variations in travel distances are relatively minor and comparable among UAVs, making the constant-time assumption a reasonable simplification.
}


\section{Conclusion}\label{sec:conclusion}
In this paper, an optimal scheduling-based framework for a multi-agent aerial 3D printing mission is addressed.
A framework is proposed for scheduling multiple UAVs to minimize the makespan of a construction mission. Potential inter-UAV collisions, task dependencies arising from geometric constraints between chunks are addressed, and adherence to battery life and material capacity limitations is ensured.
The computation time of the overall mission plan is reduced by prioritizing tasks with higher dependencies, while UAV utilization is maximized by dynamically adjusting the number of UAVs based on the mission's specific requirements.
The evaluation of the proposed framework indicates that the prioritization of the important tasks results in a reduction of the computational time at the expense of a slight increase in the mission makespan.
Additionally, the proposed agent utilization problem computes the optimal number of UAVs and their corresponding schedule in one step instead of brute-forcing, significantly reducing the total cost and improving efficacy.
Finally, the proposed framework is demonstrated in a simulated Gazebo environment where a given mission of constructing a  $2x2$ m. rectangle by dynamically selecting a team of four UAVs satisfying the material and flight time constraints while ensuring their safety.
%
\textcolor{blue}{
Future work includes hardware experimentation of aerial 3D printing using the proposed framework and the exploration of advanced algorithmic techniques, such as branch-and-price and metaheuristics, to enhance computational efficiency.}
 
    


\bibliographystyle{IEEEtran}
\bibliography{sample}

@inproceedings{al2018large,
  title={Large-scale 3D printing: the way forward},
  author={Al Jassmi, Hamad and Al Najjar, Fady and Mourad, Abdel-Hamid Ismail},
  booktitle={IOP Conference Series: Materials Science and Engineering},
  volume={324},
  pages={012088},
  year={2018},
  organization={IOP Publishing},
  doi = {10.1088/1757-899X/324/1/012088}
}

@article{bazli20233d,
  title={3D printing for remote housing: Benefits and challenges},
  author={Bazli, Milad and Ashrafi, Hamed and Rajabipour, Ali and Kutay, Cat},
  journal={Automation in Construction},
  volume={148},
  pages={104772},
  year={2023},
  publisher={Elsevier},
doi = {https://doi.org/10.1016/j.autcon.2023.104772}
}

@article{craveiroa2019additive,
  title={Additive manufacturing as an enabling technology for digital construction: A perspective on Construction 4.0},
  author={Craveiroa, Fl{\'a}vio and Duartec, Jos{\'e} Pinto and Bartoloa, Helena and Bartolod, Paulo Jorge},
  journal={Sustain. Dev},
  volume={4},
  number={6},
  year={2019}
}

@article{zhang2022aerial,
  title={Aerial additive manufacturing with multiple autonomous robots},
  author={Zhang, Ketao and Chermprayong, Pisak and Xiao, Feng and Tzoumanikas and others},
  journal={Nature},
  volume={609},
  number={7928},
  pages={709--717},
  year={2022},
  publisher={Nature Publishing Group UK London},
  doi = {https://doi.org/10.1038/s41586-022-04988-4}
}

@article{xu2022robotics,
  title={Robotics technologies aided for 3D printing in construction: A review},
  author={Xu, Zhen and Song, Tao and Guo, Shuai and Peng, Jiangtao and Zeng, Lingdong and Zhu, Mengmeng},
  journal={The International Journal of Advanced Manufacturing Technology},
  volume={118},
  number={11-12},
  pages={3559--3574},
  year={2022},
  publisher={Springer},
  doi = {10.1007/s00170-021-08067-2}
}

@article{nature_aerial_AM,
author = {Zhang, Ketao and Chermprayong, Pisak and Xiao, Feng and Tzoumanikas, Dimos et al},
issn = {14764687},
journal = {Nature},
number = {7928},
pages = {709--717},
pmid = {36131037},
publisher = {Springer US},
title = {{Aerial additive manufacturing with multiple autonomous robots}},
volume = {609},
year = {2022}
}

@article{wenchao_zhou_ground_robots,
title = {Toward Swarm Manufacturing: Architecting a Cooperative 3D Printing System},
author = {Poudel, Laxmi and Marques, Lucas Galvan and Williams, Robert Austin and Hyden, Zachary and Guerra, Pablo and Fowler, Oliver Luke and Sha, Zhenghui and Zhou, Wenchao},
issn = {15288935},
journal = {Journal of Manufacturing Science and Engineering, Transactions of the ASME},
number = {8},
pages = {1--15},
volume = {144},
year = {2022}
}

@INPROCEEDINGS{carma,
  title={On the design, modeling and control of a novel compact aerial manipulator},
  author={Wuthier, D. and Kominiak, D. and Kanellakis, C. and Andrikopoulos, G. and Fumagalli, M. and Schipper, G. and Nikolakopoulos, G.},
  booktitle={2016 24th Mediterranean Conference on Control and Automation (MED)},  
  year={2016}
  }

@book{bigM_notation,
  title={Integer programming},
  author={Wolsey, Laurence A},
  year={2020},
  publisher={John Wiley \& Sons},
  doi = {http://dx.doi.org/10.1002/9781119606475.oth1}
}

@article{3dPrintingTeamRobots,
title = {Large-scale 3D printing by a team of mobile robots},
journal = {Automation in Construction},
volume = {95},
pages = {98-106},
year = {2018},
issn = {0926-5805},
author = {Xu Zhang and Mingyang Li and Jian Hui Lim and Yiwei Weng and Yi Wei Daniel Tay and Hung Pham and Quang-Cuong Pham},
}

@inproceedings{kovac,
  title={3D printing with flying robots},
  author={Hunt, Graham and Mitzalis, Faidon and Alhinai, Talib and Hooper, Paul A and Kovac, Mirko},
  booktitle={2014 IEEE international conference on robotics and automation (ICRA)},
  pages={4493--4499},
  year={2014},
  organization={IEEE}
}

@incollection{reviewAddManAerospace,
title = {2 - Review of additive manufacturing technologies and applications in the aerospace industry},
editor = {Francis Froes and Rodney Boyer},
booktitle = {Additive Manufacturing for the Aerospace Industry},
publisher = {Elsevier},
pages = {7-31},
year = {2019},
isbn = {978-0-12-814062-8},
doi = {https://doi.org/10.1016/B978-0-12-814062-8.00002-9},
url = {https://www.sciencedirect.com/science/article/pii/B9780128140628000029},
author = {Joel C. Najmon and Sajjad Raeisi and Andres Tovar},
}

@article{stamatopoulosChunkingJINT,
  author    = {Marios-Nektarios Stamatopoulos and Avijit Banerjee and George Nikolakopoulos},
  title     = {A Decomposition and a Scheduling Framework for Enabling Aerial 3D Printing},
  journal   = {Journal of Intelligent \& Robotic Systems},
  year      = {2024},
  volume    = {110},
  number    = {2},
  pages     = {53},
  doi       = {10.1007/s10846-024-02081-8},
  url       = {https://doi.org/10.1007/s10846-024-02081-8},
  issn      = {1573-0409}
}

@article{stamatopoulos2024conflict,
  title={Conflict-free optimal motion planning for parallel aerial 3D printing using multiple UAVs},
  author={Stamatopoulos, Marios-Nektarios and Banerjee, Avijit and Nikolakopoulos, George},
  journal={Expert Systems with Applications},
  volume={246},
  pages={123201},
  year={2024},
  publisher={Elsevier}
}

@article{batteryLifeSina,
title = {Remaining Useful Battery Life Prediction for UAVs based on Machine Learning**This work has received partial funding from the European Union’s Horizon 2020 Research and Innovation Programme under the Grant Agreement No.644128, AEROWORKS},
journal = {IFAC-PapersOnLine},
volume = {50},
number = {1},
pages = {4727-4732},
year = {2017},
note = {20th IFAC World Congress},
issn = {2405-8963},
doi = {https://doi.org/10.1016/j.ifacol.2017.08.863},
url = {https://www.sciencedirect.com/science/article/pii/S2405896317313253},
author = {Sina Sharif Mansouri and Petros Karvelis and George Georgoulas and George Nikolakopoulos}
}

@inproceedings{StamatopoulosEmulation,
 author={Stamatopoulos, Marios-Nektarios and Banerjee, Avijit and Nikolakopoulos, George},
  booktitle={2024 IEEE International Conference on Robotics and Automation (ICRA)}, 
  title={On Experimental Emulation of Printability and Fleet Aware Generic Mesh Decomposition for Enabling Aerial 3D Printing}, 
  year={2024},
  volume={},
  number={},
  pages={10080-10086},
  doi={10.1109/ICRA57147.2024.10610806}
}

@INPROCEEDINGS{bjornCollisionAvoidance,
  author={Lindqvist, Björn and Sopasakis, Pantelis and Nikolakopoulos, George},
  booktitle={2021 IEEE/RSJ International Conference on Intelligent Robots and Systems (IROS)}, 
  title={A Scalable Distributed Collision Avoidance Scheme for Multi-agent UAV systems}, 
  year={2021},
  volume={},
  number={},
  pages={9212-9218},
  doi={10.1109/IROS51168.2021.9636293}}

@misc{gurobi,
  author = {{Gurobi Optimization, LLC}},
  title = {{Gurobi Optimizer Reference Manual}},
  year = 2024,
  url = "https://www.gurobi.com"
}

@INPROCEEDINGS{stamatopoulosICUAS,
  author={Stamatopoulos, Marios-Nektarios and Banerjee, Avijit and Nikolakopoulos, George},
  booktitle={2024 International Conference on Unmanned Aircraft Systems (ICUAS)}, 
  title={Collaborative Aerial 3D Printing: Leveraging UAV Flexibility and Mesh Decomposition for Aerial Swarm-Based Construction}, 
  year={2024},
  volume={},
  number={},
  pages={45-52},
  doi={10.1109/ICUAS60882.2024.10557090}}

@article{review2019jobshopIndustry4.0,
  author    = {Jian Zhang and Guohua Ding and Yan Zou and others},
  title     = {Review of job shop scheduling research and its new perspectives under Industry 4.0},
  journal   = {Journal of Intelligent Manufacturing},
  volume    = {30},
  pages     = {1809--1830},
  year      = {2019},
  doi       = {10.1007/s10845-017-1350-2},
  url       = {https://doi.org/10.1007/s10845-017-1350-2}
}

@article{surveyJobScheduling,
title = {A survey of job shop scheduling problem: The types and models},
journal = {Computers \& Operations Research},
volume = {142},
pages = {105731},
year = {2022},
issn = {0305-0548},
doi = {https://doi.org/10.1016/j.cor.2022.105731},
url = {https://www.sciencedirect.com/science/article/pii/S0305054822000338},
author = {Hegen Xiong and Shuangyuan Shi and Danni Ren and Jinjin Hu},
}

@article{milpJobShopMobileRobots,
title = {A novel MILP model for job shop scheduling problem with mobile robots},
journal = {Robotics and Computer-Integrated Manufacturing},
volume = {81},
pages = {102506},
year = {2023},
issn = {0736-5845},
doi = {https://doi.org/10.1016/j.rcim.2022.102506},
url = {https://www.sciencedirect.com/science/article/pii/S0736584522001880},
author = {You-Jie Yao and Qi-Hao Liu and Xin-Yu Li and Liang Gao},
}

@article{flexibleJobShopmilpEnergy,
title = {MILP models for energy-aware flexible job shop scheduling problem},
journal = {Journal of Cleaner Production},
volume = {210},
pages = {710-723},
year = {2019},
issn = {0959-6526},
doi = {https://doi.org/10.1016/j.jclepro.2018.11.021},
url = {https://www.sciencedirect.com/science/article/pii/S0959652618334218},
author = {Leilei Meng and Chaoyong Zhang and Xinyu Shao and Yaping Ren},
}

@article{taxonomyTaskAllocation,
title = {A taxonomy for task allocation problems with temporal and ordering constraints},
journal = {Robotics and Autonomous Systems},
volume = {90},
pages = {55-70},
year = {2017},
note = {Special Issue on New Research Frontiers for Intelligent Autonomous Systems},
issn = {0921-8890},
doi = {https://doi.org/10.1016/j.robot.2016.10.008},
url = {https://www.sciencedirect.com/science/article/pii/S0921889016306157},
author = {Ernesto Nunes and Marie Manner and Hakim Mitiche and Maria Gini}
}

@InProceedings{JSCwithAGV,
author="Geitz, Marc
and Grozea, Cristian
and Steigerwald, Wolfgang
and St{\"o}hr, Robin
and Wolf, Armin",
editor="Schaus, Pierre",
title="Solving the Extended Job Shop Scheduling Problem with AGVs -- Classical and Quantum Approaches",
booktitle="Integration of Constraint Programming, Artificial Intelligence, and Operations Research",
year="2022",
publisher="Springer International Publishing",
address="Cham",
pages="120--137",
isbn="978-3-031-08011-1"
}

@article{JSPdeadlcokMovemConsiderations,
title = {Novel robotic job-shop scheduling models with deadlock and robot movement considerations},
journal = {Transportation Research Part E: Logistics and Transportation Review},
volume = {149},
pages = {102273},
year = {2021},
issn = {1366-5545},
doi = {https://doi.org/10.1016/j.tre.2021.102273},
url = {https://www.sciencedirect.com/science/article/pii/S1366554521000491},
author = {Yige Sun and Sai-Ho Chung and Xin Wen and Hoi-Lam Ma},
}

@article{schedulingAircraftSoftPrecedence,
author = {Tereshchuk, Veniamin and Bykov, Nikolay and Pedigo, Samuel and Devasia, Santosh and Banerjee, Ashis G.},
doi = {10.1016/j.rcim.2021.102154},
issn = {07365845},
journal = {Robotics and Computer-Integrated Manufacturing},
number = {March},
pages = {102154},
publisher = {Elsevier Ltd},
title = {A scheduling method for multi-robot assembly of aircraft structures with soft task precedence constraints},
url = {https://doi.org/10.1016/j.rcim.2021.102154},
volume = {71},
year = {2021}
}

@article{spacecraftmanufacBalanced,
author = {Liu, Shaorui and Shen, Jianxin and Tian, Wei and Lin, Jiamei and Li, Pengcheng and Li, Bo},
doi = {10.1016/j.robot.2022.104289},
issn = {09218890},
journal = {Robotics and Autonomous Systems},
title = {Balanced task allocation and collision-free scheduling of multi-robot systems in large spacecraft structure manufacturing},
volume = {159},
year = {2023}
}

@INPROCEEDINGS{scheduling3DPrinters,
  author={Kim, Jun and Park, Sang-Soo and Kim, Hyun-Jung},
  booktitle={2017 13th IEEE Conference on Automation Science and Engineering (CASE)}, 
  title={Scheduling 3D printers with multiple printing alternatives}, 
  year={2017},
  volume={},
  number={},
  pages={488-493},
  doi={10.1109/COASE.2017.8256151}}

@article{wenchaoZhouScheduling,
author = {Poudel, Laxmi and Zhou, Wenchao and Sha, Zhenghui},
doi = {10.1115/1.4050380},
issn = {10500472},
journal = {Journal of Mechanical Design},
number = {7},
pages = {1--12},
title = {{Resource-constrained scheduling for multi-robot cooperative three-dimensional printing}},
volume = {143},
year = {2021}
}

@article{addManIndustry4.0,
author = {Haleem, Abid and Javaid, Mohd},
title = {Additive Manufacturing Applications in Industry 4.0: A Review},
journal = {Journal of Industrial Integration and Management},
volume = {04},
number = {04},
pages = {1930001},
year = {2019},
doi = {10.1142/S2424862219300011},
URL = { https://doi.org/10.1142/S2424862219300011
},
eprint = {https://doi.org/10.1142/S2424862219300011}
}

@incollection{addManAutomotive,
  title={Additive manufacturing for the automotive industry},
  author={Vasco, Joel C},
  booktitle={Additive Manufacturing},
  pages={505--530},
  year={2021},
  publisher={Elsevier}
}

@Article{addManCosntruction,
AUTHOR = {Tuvayanond, Wiput and Prasittisopin, Lapyote},
TITLE = {Design for Manufacture and Assembly of Digital Fabrication and Additive Manufacturing in Construction: A Review},
JOURNAL = {Buildings},
VOLUME = {13},
YEAR = {2023},
NUMBER = {2},
ARTICLE-NUMBER = {429},
URL = {https://www.mdpi.com/2075-5309/13/2/429},
ISSN = {2075-5309},
DOI = {10.3390/buildings13020429}
}

@Article{addManConcreteCosntruction,
AUTHOR = {Khajavi, Siavash H. and Tetik, Müge and Mohite, Ashish and Peltokorpi, Antti and Li, Mingyang and Weng, Yiwei and Holmström, Jan},
TITLE = {Additive Manufacturing in the Construction Industry: The Comparative Competitiveness of 3D Concrete Printing},
JOURNAL = {Applied Sciences},
VOLUME = {11},
YEAR = {2021},
NUMBER = {9},
ARTICLE-NUMBER = {3865},
URL = {https://www.mdpi.com/2076-3417/11/9/3865},
ISSN = {2076-3417},
DOI = {10.3390/app11093865}
}

@article{addManufMobileRobotsChallenges,
title = {Additive Manufacturing using mobile robots: Opportunities and challenges for building construction},
journal = {Cement and Concrete Research},
volume = {158},
pages = {106772},
year = {2022},
issn = {0008-8846},
doi = {https://doi.org/10.1016/j.cemconres.2022.106772},
url = {https://www.sciencedirect.com/science/article/pii/S0008884622000631},
author = {Kathrin Dörfler and Gido Dielemans and Lukas Lachmayer and Tobias Recker and Annika Raatz and Dirk Lowke and Markus Gerke},
}

@ARTICLE{TASEflexibjeJobShop,
  author={Cao, ZhengCai and Lin, ChengRan and Zhou, MengChu},
  journal={IEEE Transactions on Automation Science and Engineering}, 
  title={A Knowledge-Based Cuckoo Search Algorithm to Schedule a Flexible Job Shop With Sequencing Flexibility}, 
  year={2021},
  volume={18},
  number={1},
  pages={56-69},
  doi={10.1109/TASE.2019.2945717}}

@ARTICLE{TASEprefabConstruction,
  author={Zhu, Aiyu and Dai, Tianhong and Xu, Gangyan and Pauwels, Pieter and De Vries, Bauke and Fang, Meng},
  journal={IEEE Transactions on Automation Science and Engineering}, 
  title={Deep Reinforcement Learning for Real-Time Assembly Planning in Robot-Based Prefabricated Construction}, 
  year={2023},
  volume={20},
  number={3},
  pages={1515-1526},
  doi={10.1109/TASE.2023.3236805}}

@ARTICLE{TASEpetriNetsScheduling,
  author={Casalino, Andrea and Zanchettin, Andrea Maria and Piroddi, Luigi and Rocco, Paolo},
  journal={IEEE Transactions on Automation Science and Engineering}, 
  title={{Optimal Scheduling of Human–Robot Collaborative Assembly Operations With Time Petri Nets}}, 
  year={2021},
  volume={18},
  number={1},
  pages={70-84},
  doi={10.1109/TASE.2019.2932150}}

@Article{foamExpansion,
AUTHOR = {Li, Xiaolong and Peng, Cen and Ao, Yanna and Hao, Meimei and Zhong, Yanhui and Zhang, Bei},
TITLE = {Impact of Composition Ratio on the Expansion Behavior of Polyurethane Grout},
JOURNAL = {Materials},
VOLUME = {17},
YEAR = {2024},
NUMBER = {8},
ARTICLE-NUMBER = {1835},
URL = {https://www.mdpi.com/1996-1944/17/8/1835},
PubMedID = {38673191},
ISSN = {1996-1944},
}

@article{additiveManConstruction2,
author = {Pacillo, Gerardo Arcangelo and Ranocchiai, Giovanna and Loccarini, Federica and Fagone, Mario},
title = {Additive manufacturing in construction: A review on technologies, processes, materials, and their applications of 3D and 4D printing},
journal = {Material Design \& Processing Communications},
volume = {3},
number = {5},
pages = {e253},
doi = {https://doi.org/10.1002/mdp2.253}
}

@book{mehaffy2017design,
  title={{Design for a living planet: Settlement, science, \& the human future}},
  author={Mehaffy, Michael and Salingaros, Nikos A},
  year={2017},
  publisher={Sustasis Press}
}

@article{
scienceRobAAMsurvey,
author = {Yusuf Furkan Kaya  and Lachlan Orr  and Basaran Bahadir Kocer  and Vijay Pawar  and Robert Stuart-Smith  and Mirko Kovač },
title = {Aerial additive manufacturing: Toward on-site building construction with aerial robots},
journal = {Science Robotics},
volume = {10},
number = {101},
pages = {eado6251},
year = {2025},
doi = {10.1126/scirobotics.ado6251},
URL = {https://www.science.org/doi/abs/10.1126/scirobotics.ado6251},
}

@article{AUTCONstamatopoulos2025experiment,
title = {Fully autonomous chunk-based aerial additive manufacturing with Offset-free Predictive Control},
journal = {Automation in Construction},
volume = {178},
pages = {106361},
year = {2025},
issn = {0926-5805},
doi = {https://doi.org/10.1016/j.autcon.2025.106361},
url = {https://www.sciencedirect.com/science/article/pii/S0926580525004017},
author = {Marios-Nektarios Stamatopoulos and Jakub Haluška and Elias Small and Jude Marroush and Avijit Banerjee and George Nikolakopoulos},
}

@article{ETHAerialMasonry,
author = {Augugliaro, Frederico and Lupashin, Sergei and Hamer, Michael and Male, Cason and Hehn, Markus and Mueller, Mark W. and Willmann, Jan Sebastian and Gramazio, Fabio and Kohler, Matthias and D'Andrea, Raffaello},
doi = {10.1109/MCS.2014.2320359},
issn = {1066033X},
journal = {IEEE Control Systems},
number = {4},
pages = {46--64},
title = {{The flight assembled architecture installation: Cooperative construction with flying machines}},
volume = {34},
year = {2014}
}

@article{metaheuristicAerialMasonry,
author = {Elkhapery, Basel and P{\v{e}}ni{\v{c}}ka, Robert and N{\v{e}}mec, Michal and Siddiqui, Mohsin},
doi = {10.1016/j.autcon.2023.104908},
issn = {09265805},
journal = {Automation in Construction},
number = {April},
pages = {104908},
publisher = {Elsevier B.V.},
title = {{Metaheuristic planner for cooperative multi-agent wall construction with UAVs}},
url = {https://doi.org/10.1016/j.autcon.2023.104908},
volume = {152},
year = {2023}
}

@INPROCEEDINGS{youwasps,
  author={Sustarevas, Julius and Benjamin Tan, K. X. and Gerber, David and Stuart-Smith, Robert and Pawar, Vijay M.},
  booktitle={2019 IEEE/RSJ International Conference on Intelligent Robots and Systems (IROS)}, 
  title={YouWasps: Towards Autonomous Multi-Robot Mobile Deposition for Construction}, 
  year={2019},
  volume={},
  number={},
  pages={2320-2327},
  doi={10.1109/IROS40897.2019.8967766}}

@article{stamatopoulos2025Masonry,
  title={Autonomous Reactive Masonry Construction using Collaborative Heterogeneous Aerial Robots with Experimental Demonstration},
  author={Stamatopoulos, Marios-Nektarios and Small, Elias and Velhal, Shridhar and Banerjee, Avijit and Nikolakopoulos, George},
  journal={arXiv preprint arXiv:2510.15114},
  year={2025}
}

\end{document}